\pdfoutput=1
\documentclass[11pt]{article}

\usepackage[preprint]{acl}
\usepackage{hyperref}
\usepackage{url}
\usepackage{amsmath}
\usepackage{amssymb}
\usepackage{amsfonts}
\usepackage{booktabs}
\usepackage{multirow}
\usepackage{graphicx}
\usepackage{makecell}
\usepackage{subcaption}
\usepackage{enumitem}
\usepackage[table]{xcolor}

\newcommand{\secbg}{\rowcolor{gray!15}}
\definecolor{oursbg}{RGB}{235,243,252}
\newcommand{\comp}[3]{%
  \begin{tabular}[c]{@{}c@{}}
    #1\,{\tiny(#2)}\\[-0.4ex]
    {\tiny\color{gray}#3}
  \end{tabular}}

\newcommand{\ours}[3]{%
  \cellcolor{oursbg}%
  \begin{tabular}[c]{@{}c@{}}
    #1\,{\tiny(#2)}\\[-0.4ex]
    {\tiny\color{gray}#3}
  \end{tabular}}

\newcommand{\methodcite}[2]{%
\makecell[l]{#1\\[-1pt]{\scriptsize\citep{#2}}}%
}

\usepackage{times}
\usepackage{latexsym}

\usepackage[T1]{fontenc}

\usepackage[utf8]{inputenc}

\usepackage{microtype}

\usepackage{inconsolata}

\usepackage{graphicx}

\title{The Butterfly Effect of Language-Specific Components in Multilingual Vision-Language Models}

\author{
  \textbf{Weihua Zheng\textsuperscript{1,2}},
  \textbf{Rui Cao\textsuperscript{1}},
  \textbf{Xiyan Tao\textsuperscript{2}},
  \textbf{Zhengyuan Liu\textsuperscript{2}},
  \textbf{Nancy F. Chen\textsuperscript{2}},
  \textbf{Roy Ka-Wei Lee\textsuperscript{3}}
\\
  \textsuperscript{1}Singapore University of Technology and Design \\
  \textsuperscript{2}Agency for Science, Technology and Research \\
  \textsuperscript{3}University of British Columbia
\\
  \texttt{zhengw@a-star.edu.sg}
}

\newsavebox{\myfirsttablebox}

\begin{document}
\maketitle

\begin{abstract}

Multilingual vision-language models exhibit cross-lingual disparities in hallucination, even when the image and question semantics remain unchanged. We hypothesize that small language-dependent deviations in hidden states accumulate during autoregressive decoding, progressively weakening visual grounding. Motivated by this butterfly effect, we propose Multilingual Visually Protected Language-Residual Steering (MVP-Steer), a training-free method that applies lightweight hidden-state corrections throughout generation. We decompose hidden representations into language centroids, cross-lingually shared semantics, visually sensitive components, and non-shared residuals. During decoding, we selectively attenuate non-shared residuals while protecting language identity, shared semantics, and visual information. Across three VLMs, ten languages, and three benchmarks, MVP-Steer reduces hallucination by up to 13.8\% while preserving generation quality, achieving a better trade-off than existing training-free methods. Further analysis shows that frequent small corrections achieve a more favorable balance between hallucination reduction and visual coverage than sparse interventions under matched perturbation energy, supporting the proposed butterfly-effect mechanism.

\end{abstract}

\section{Introduction}

Vision-language models (VLMs) have achieved strong performance in visual question answering, image description, and multimodal reasoning,
and are increasingly extended to multilingual settings~\citep{maaz2024palo,geigle2024mblip}.
However, visual hallucination, generating content unsupported by the input image, remains a persistent limitation.
In multilingual VLMs, hallucination rates can vary substantially across languages even when the image and query semantics are held fixed~\citep{qu2026mitigating,ye2025claim}.
Mitigating it in multilingual VLMs requires preserving visual grounding while supporting generation in the requested language.

Existing methods address hallucination through
decoding-time intervention or hallucination-aware training in general VLMs \citep{leng2024mitigating,huang2024opera,liu2024mitigating,yu2024rlhfv}, and through cross-lingual preference optimization or attention intervention in multilingual VLMs \citep{qu2026mitigating,ye2025claim}.
However, they largely overlook the cross-lingual structure of hidden representations.
Intervening on this structure is challenging because components outside the shared cross-lingual subspace may also encode visual evidence, and naively suppressing all non-shared components may compromise visual grounding.

To investigate how this structure can inform hallucination mitigation, we examine intermediate hidden representations, where visual evidence and language context jointly shape generation.
Prior work characterizes multilingual representations as comprising a cross-lingually shared semantic structure and language-specific components \citep{pires2019multilingual,libovicky2020language,chang2022geometry,tezuka2025transfer}.
Language-dependent variation supports language control but may also contain visually grounded information. We hypothesize that small cross-lingual differences can accumulate during autoregressive generation and contribute to hallucination disparities. This motivates selective, repeated correction of non-shared residuals while preserving shared semantics, language control, and visual evidence.

We propose \textit{\textbf{M}ultilingual \textbf{V}isually \textbf{P}rotected Language-Residual Steering} (MVP-Steer), an inference-time method that leaves model weights unchanged.
MVP-Steer decomposes each hidden state into a language centroid that carries the language-control signal, a shared semantic component, a visually sensitive component, and a remaining non-shared residual. During generation, it attenuates the residual, with partial protection of the visually sensitive component, at every decoding step with an update scaled to the norm of the current hidden state.
We evaluate MVP-Steer on AMBER~\citep{wang2023amber}, LLaVA-Bench~\citep{liu2024improved, liu2023visual}, and Object HalBench~\citep{yu2024rlhfv} using Gemma-3-12B-IT~\citep{team2025gemma}, InternVL3.5-14B-IT~\citep{wang2025internvl3}, and Qwen3-VL-8B-IT~\citep{bai2025qwen3} across ten languages. MVP-Steer reduces AMBER CHAIR by up to $13.8\%$ while preserving generation quality, achieving a better trade-off than existing training-free methods. At the same total perturbation energy, ablations favor frequent small corrections over sparse corrections: they reduce hallucinations more effectively without sacrificing object coverage. This finding is consistent with the butterfly-effect hypothesis.

Our contributions are: (i) We characterize cross-lingual hidden-state structure and provide evidence consistent with autoregressive amplification of language-dependent deviations. (ii) We propose MVP-Steer, which retains language centroids and shared semantic components while attenuating non-shared variation with partial protection for visually sensitive components. (iii) Across ten languages, three VLMs, and three benchmarks, we demonstrate reduced language-averaged object hallucination with comparable average generation quality.

\section{Related Work}
\label{sec:related}

\paragraph{Hallucination in multilingual vision-language models.} Visual hallucination evaluation has evolved from measuring unsupported objects in captions with CHAIR~\citep{rohrbach2018object} to assessing broader hallucination types with AMBER~\citep{wang2023amber}. Decoding analyses show that early errors can propagate across generation steps~\citep{zhong2024investigating} as visual information gradually loses influence~\citep{li2025hidden}. These dynamics are particularly relevant in multilingual VLMs, where hallucination rates vary across languages~\citep{schneider2024m5} and non-English queries can amplify hallucination~\citep{qu2026mitigating}. These findings shift attention from cross-lingual differences to how language-dependent representations affect visual grounding during generation. \paragraph{Hallucination mitigation and representation steering.} Training-based methods improve visual grounding through robust instruction tuning~\citep{liu2024mitigating}, reinforcement learning~\citep{sun2024aligning,yu2024rlhfv}, or preference optimization~\citep{zhao2023beyond}, but require additional supervision and parameter updates. Training-free methods instead intervene during inference on decoding distributions, attention patterns, or internal representations~\citep{wang2024mitigating,favero2024multimodal}. At the decoding level, VCD~\citep{leng2024mitigating} contrasts output distributions from original and distorted images, and OPERA~\citep{huang2024opera} penalizes attention that concentrates on a few summary tokens during beam-search decoding. At the attention level, AdaIAT~\citep{zhong2026adaiat} adaptively increases attention to previously generated text tokens, and CLAIM~\citep{ye2025claim} intervenes on attention heads associated with cross-lingual variation to mitigate multilingual hallucination. However, these approaches do not explicitly separate cross-lingually shared structure from language-dependent variation or protect the visual information in the non-shared space. \paragraph{Shared and language-specific structure in multilingual representations.} In multilingual models, cross-lingually shared semantic structure coexists with language-specific variation~\citep{pires2019multilingual,libovicky2020language,chang2022geometry,tezuka2025transfer}. Studies of decoder-only LLMs show that intermediate layers organize concepts in partially shared spaces, while language identity concentrates in a small set of directions or neurons~\citep{wendler2024llamas,wu2025semantic,tang2024language}. This geometry has motivated methods that remove language-specific information through PCA-based projection~\citep{yang2021simple} and language-subspace manipulation~\citep{zhao2024lens}. However, these text-based methods overlook visual evidence that may occupy the same non-shared space in VLMs. MVP-Steer protects visually sensitive directions and attenuates only the remaining non-shared residual during generation.

\section{Motivating Observations}
\label{sec:Motivating Observations}

Multilingual VLMs hallucinate at markedly different rates across languages despite identical images and semantically equivalent queries. We trace the hallucination disparity across languages through four representational observations: shared semantic structure, the effects of centroid shifts on output language, visual-attention consistency under controlled content, and cross-lingual deviations during autoregressive decoding. Each of them informs one part of MVP-Steer.

\paragraph{Setup.} For Observation~1, we analyze InternVL3.5-14B-IT and Qwen3-VL-8B-IT on 4,146 aligned image--description instances in ten target languages plus English. Observation~2 conducts causal centroid interventions on InternVL3.5-14B-IT for Japanese, Korean, and Vietnamese, using 200 AMBER samples per language. Observations~3 and~4 analyze the baseline InternVL3.5-14B-IT using 100 held-out AMBER images with semantically equivalent prompts in the ten target languages. Details are provided in App.~\ref{app:geometry_analysis}, \ref{app:centroid_intervention}, and~\ref{app:visual_grounding} for Observations~1 to~3, and in App.~\ref{app:attention_free_running} and~\ref{app:butterfly_empirical} for Observation~4.

\paragraph{Observation 1: Shared semantic structure coexists with language-dependent centroid offsets.} Motivated by findings in multilingual language models that shared semantic structure coexists with language-dependent variation~\citep{libovicky2020language,chang2022geometry}, we examine the cross-lingual structure of multilingual VLM representations. We perform bidirectional cross-lingual retrieval across 45 target-language pairs, retrieving the semantically matched instance in one language from its representation in another. Averaged across pairs, Recall@1 reaches $0.9866$ for InternVL3.5-14B-IT and $0.9767$ for Qwen3-VL-8B-IT, indicating correspondence between matched instances.
Representational similarity analysis (RSA) yields averages of $0.7615$ and $0.7740$ on InternVL and Qwen, respectively, indicating broadly similar semantic geometry across languages.

Similar geometry, however, does not imply the same location in hidden space.  Following \citet{libovicky2020language}, we subtract each language's centroid, defined as its mean representation across semantic instances, to remove language-dependent mean offsets. This increases the semantic margin, which is the difference between mean cosine similarities of matched and mismatched cross-lingual pairs, from 0.18 to 0.66 for InternVL and from 0.09 to 0.69 for Qwen. Mismatched cosine similarity simultaneously decreases from 0.65 to 0.0006 and from 0.81 to 0.001, respectively.

Thus, shared semantic structure coexists with language-dependent centroid offsets, but improved alignment after centering does not show whether the centroid should be modified during generation.

\paragraph{Observation 2: Centroid shifts alter output language and object coverage. }
We test the causal role of the centroid by shifting textual hidden states along the centroid-difference direction toward English.
On InternVL3.5-14B-IT,
an intermediate-layer perturbation of 0.3\% of the hidden-state norm switches all Japanese, Korean, and Vietnamese responses to English.
These responses still mention approximately $54\%$ of the annotated objects in the images used for the observation experiment. This shows that centroid differences provide a causally effective language-control signal.

We then apply the opposite centroid direction in upper layers.
This restores the target language in $70\%$ to $100\%$ of responses, but does not recover the descriptive capability affected by the earlier intervention: average object Cover drops from 0.52 to 0.08. Restoring the output language alone is therefore insufficient to undo the effects of an earlier centroid shift.
These results motivate preserving the language centroid during inference and applying corrections to the centered representation obtained by subtracting it.

\paragraph{
Observation 3: Visual attention is largely consistent across languages under controlled content.
}
We compare cross-lingual visual attention immediately before generation and throughout descriptions with controlled content.
At the final prompt token, visual attention distributions have an average cosine similarity of
$0.9514$ across all 45 target-language pairs, indicating highly similar attention patterns over images.
To examine attention during decoding while controlling description content, we teacher-force a canonical English description and its ten translations, supplying reference prefixes instead of self-generated tokens.
After aggregating attention over semantic units within each language, teacher forcing yields a cross-lingual visual-attention cosine similarity of $0.9603$ (Appendix~\ref{app:visual_grounding}). These findings suggest that visual attention is largely consistent across languages when semantic content is controlled.
However, teacher forcing excludes the influence of self-generated tokens. We next examine whether this consistency persists when each language generates its own response and subsequent decoding conditions on its previously generated tokens.

\paragraph{Observation 4: Unconstrained generation amplifies small deviations across languages.}
We compare free-running decoding, which conditions on self-generated tokens, with teacher forcing to examine the autoregressive-amplification hypothesis.
Cross-lingual visual-attention cosine similarity is $0.9492$ under free-running decoding and $0.9603$ under teacher forcing. Controlling description content therefore reduces attention dissimilarity, defined as one minus cosine similarity, by $22.0\%$ (Appendix~\ref{app:attention_free_running}).

We measure cosine distance between target-language and English hidden states, each centered by subtracting its language centroid, at matched relative response positions. From early to late generation, this distance increases more under unconstrained generation than under teacher forcing, by $0.0831$ on average. This holds in all ten languages and in $71.1\%$ of image and language pairs.
Object-hallucination mention rates rise from $1.43\%$ to $6.55\%$ across three response segments, with higher late-stage rates in every language (Appendix~\ref{app:butterfly_empirical}).
These findings are consistent with the autoregressive-amplification hypothesis and motivate repeated corrections during decoding.

\paragraph{From observations to representation decomposition.}

These observations motivate selective correction of non-shared variation while preserving language control and visual grounding.
Observation~1 shows that a shared semantic geometry coexists with
language-dependent offsets, motivating an explicit separation between the language centroid and centered semantic variation.
Observation~2 identifies the centroid as a causal language-control signal that should be retained.
Observation~3 indicates that visual grounding is largely consistent across languages when semantic content is controlled.
However, the shared semantic subspace is estimated from cross-lingual textual correspondence and has limited rank, so it may omit useful visual information. We therefore also protect visually grounded directions outside it.
Observation~4 shows that small deviations across languages grow during unconstrained generation, which motivates small corrections at every decoding step.
These findings lead naturally to a representation decomposition that
preserves the language centroid, cross-lingually shared semantics, and
visually supported non-shared components, while isolating the remaining non-shared residual for repeated correction during decoding.

\begin{table*}[t]
\centering
\captionsetup{skip=5pt}
\scriptsize

\setlength{\tabcolsep}{2.5pt}
\renewcommand{\arraystretch}{1.35}

\resizebox{\textwidth}{!}{%
\begin{tabular}{@{}lcccccccc@{}}
\toprule
\multirow{2}{*}{\textbf{Model}}
& \multicolumn{4}{c}{\textbf{AMBER}}
& \textbf{LLaVA-Bench}
& \multicolumn{3}{c}{\textbf{Object HalBench}} \\
\cmidrule(lr){2-5}\cmidrule(lr){6-6}\cmidrule(l){7-9}
& \textbf{Cover}$\uparrow$
& \textbf{CHAIR}$\downarrow$
& \textbf{Hal}$\downarrow$
& \textbf{Cog}$\downarrow$
& \textbf{Overall}$\uparrow$
& \textbf{CHAIR$_s$}$\downarrow$
& \textbf{CHAIR$_i$}$\downarrow$
& \textbf{Recall}$\uparrow$ \\
\specialrule{\lightrulewidth}{\aboverulesep}{0pt}

\secbg\multicolumn{9}{@{}c@{}}{\rule{0pt}{2.4ex}\textbf{Baselines}} \\
\specialrule{\lightrulewidth}{0pt}{\belowrulesep}
Gemma-3-12B-IT
& 56.53 & 5.99 & 19.07 & 4.24 & 67.81 & 14.33 & 7.64 & 59.71 \\
Qwen3-VL-8B-IT
& 58.74 & 6.01 & 20.42 & 4.63 & 58.34 & 9.85 & 5.84 & 48.70 \\
InternVL3.5-14B-IT
& 48.47 & 4.52 & 12.98 & 2.84 & 54.78 & 20.47 & 10.95 & 58.31 \\
InternVL3.5-14B-IT$^\dagger$
& 50.90 & 4.33 & 12.83 & 2.79 & 58.95 & 17.36 & 9.53 & 58.40 \\
\specialrule{\lightrulewidth}{\aboverulesep}{0pt}

\secbg\multicolumn{9}{@{}c@{}}{\rule{0pt}{2.4ex}
\textbf{Existing Hallucination Mitigation Methods}} \\
\specialrule{\lightrulewidth}{0pt}{\belowrulesep}

\methodcite{InternVL3.5-14B-IT$^\ddagger$ + VCD}{leng2024mitigating}
& \comp{48.61}{$+0.04$}{$[-0.48,0.61]$}
& \comp{4.68}{$-5.00$}{$[-17.66,5.14]$}
& \comp{13.77}{$-8.29$}{$[-19.47,1.39]$}
& \comp{3.03}{$-8.26$}{$[-18.87,0.99]$}
& \comp{53.47}{$-2.39$}{$[-5.22,0.49]$}
& \comp{22.05}{$-8.33$}{$[-15.26,-2.06]$}
& \comp{11.82}{$-8.06$}{$[-12.16,-3.62]$}
& \comp{60.42}{$+2.67$}{$[1.30,4.02]$} \\

\methodcite{InternVL3.5-14B-IT + AdaIAT}{zhong2026adaiat}
& \comp{47.44}{$-2.13$}{$[-3.35,-1.22]$}
& \comp{4.22}{$+6.64$}{$[1.65,11.60]$}
& \comp{12.18}{$+6.16$}{$[1.72,10.43]$}
& \comp{2.66}{$+6.34$}{$[2.00,10.57]$}
& \comp{49.72}{$-9.24$}{$[-10.84,-7.40]$}
& \comp{20.08}{$+1.89$}{$[-3.42,7.39]$}
& \comp{10.70}{$+2.28$}{$[-2.15,6.93]$}
& \comp{57.69}{$-1.07$}{$[-2.43,0.35]$} \\

\methodcite{InternVL3.5-14B-IT + CLAIM}{ye2025claim}
& \comp{47.96}{$-1.05$}{$[-3.78,0.89]$}
& \comp{4.27}{$+5.46$}{$[-2.53,14.00]$}
& \comp{12.36}{$+4.78$}{$[-2.85,12.89]$}
& \comp{2.72}{$+4.37$}{$[-3.30,12.80]$}
& \comp{52.65}{$-3.89$}{$[-6.32,-1.67]$}
& \comp{20.16}{$+1.53$}{$[-3.72,7.07]$}
& \comp{10.85}{$+0.91$}{$[-3.80,5.32]$}
& \comp{57.31}{$-1.71$}{$[-3.55,0.16]$} \\

\methodcite{InternVL3.5-14B-IT$^\dagger$ + OPERA}{huang2024opera}
& \comp{50.82}{$-0.14$}{$[-1.19,1.48]$}
& \comp{4.44}{$-2.48$}{$[-7.82,2.65]$}
& \comp{13.06}{$-1.79$}{$[-5.04,1.66]$}
& \comp{2.86}{$-2.71$}{$[-6.14,0.88]$}
& \comp{57.23}{$-2.93$}{$[-4.98,-0.86]$}
& \comp{17.86}{$-2.88$}{$[-9.77,3.06]$}
& \comp{9.78}{$-2.62$}{$[-8.60,3.00]$}
& \comp{58.15}{$-0.43$}{$[-1.28,0.39]$} \\
\specialrule{\lightrulewidth}{\aboverulesep}{0pt}

\secbg\multicolumn{9}{@{}c@{}}{\rule{0pt}{2.4ex}\textbf{Ours}} \\
\specialrule{\lightrulewidth}{0pt}{\belowrulesep}

Gemma-3-12B-IT + MVP-Steer
& \ours{56.44}{$-0.17$}{$[-0.70,0.39]$}
& \ours{5.46}{$+8.84$}{$[5.21,12.66]$}
& \ours{17.56}{$+7.90$}{$[4.14,11.86]$}
& \ours{3.87}{$+8.65$}{$[5.32,12.25]$}
& \ours{68.21}{$+0.59$}{$[-0.39,1.68]$}
& \ours{13.17}{$+8.06$}{$[3.25,12.62]$}
& \ours{6.98}{$+8.55$}{$[4.41,12.61]$}
& \ours{60.23}{$+0.87$}{$[0.04,1.65]$} \\

Qwen3-VL-8B-IT + MVP-Steer
& \ours{58.64}{$-0.18$}{$[-0.68,0.33]$}
& \ours{5.59}{$+7.00$}{$[4.73,9.30]$}
& \ours{19.19}{$+6.00$}{$[3.25,8.63]$}
& \ours{4.30}{$+6.94$}{$[4.05,9.75]$}
& \ours{58.93}{$+1.01$}{$[-1.61,3.52]$}
& \ours{9.09}{$+7.68$}{$[-2.89,18.67]$}
& \ours{5.52}{$+5.43$}{$[-4.65,15.54]$}
& \ours{49.39}{$+1.41$}{$[0.51,2.42]$} \\

InternVL3.5-14B-IT + MVP-Steer
& \ours{48.44}{$-0.06$}{$[-0.93,1.00]$}
& \ours{3.90}{$+13.77$}{$[8.86,19.31]$}
& \ours{11.34}{$+12.60$}{$[8.46,16.47]$}
& \ours{2.51}{$+11.79$}{$[8.20,15.23]$}
& \ours{55.41}{$+1.14$}{$[-0.44,2.73]$}
& \ours{18.45}{$+9.87$}{$[5.01,13.92]$}
& \ours{10.00}{$+8.68$}{$[3.54,13.13]$}
& \ours{59.34}{$+1.77$}{$[0.33,3.20]$} \\
\bottomrule
\end{tabular}
}
\caption{
Main results averaged across ten languages.
$\downarrow$ and $\uparrow$ indicate lower and higher values are better, respectively.
Numbers in parentheses denote relative improvement (\%) over the corresponding baseline;
for lower-is-better metrics, positive percentages indicate relative reductions.
Brackets below each entry give the 95\% confidence interval of the relative improvement (App.~\ref{app:ci}).
AdaIAT and CLAIM are compared against the standard InternVL3.5-14B-IT baseline,
whereas OPERA is compared against the InternVL3.5-14B-IT beam-search baseline
with beam size 5 (marked with $^\dagger$).
VCD is evaluated against its paired InternVL3.5-14B-IT rerun (marked with $^\ddagger$),
whose averaged baseline scores differ slightly from the standard run. App.~\ref{app:Per-Language Results of MVP-Steer} reports the full per-language results of MVP-Steer across all ten languages for each backbone model.
}
\label{tab:main_results}
\end{table*}

\section{Method}
\label{sec:method}

We propose MVP-Steer, a weight-update-free method that implements the decomposition motivated in Section~\ref{sec:Motivating Observations} through repeated token-level interventions on hidden representations. It comprises two stages. During offline fitting, we estimate phase- and layer-specific multilingual shared subspaces from aligned multilingual image–text data and identify visually sensitive directions through counterfactual image replacement. During inference, norm-controlled updates selectively  attenuate the non-shared residual while preserving the language centroid, shared semantics, and visually sensitive directions. The model parameters, visual encoder, and image tokens remain unchanged.

\subsection{Phase-Specific Representation Collection}
\label{sec:method_views}

Let $\ell \in \mathcal{L}$ denote a language, $i$ an aligned multilingual sample, $p \in \{\text{pre}, \text{dec}\}$ the prefill or decoding phase, and $k \in \mathcal{K}$ an intervened decoder layer of the language backbone of the VLM, with hidden dimension $d$. For each sample, language, phase, and layer, we collect one representation $\mathbf{h}^{(k)}_{i,\ell,p} \in \mathbb{R}^d$ from the residual stream output of that layer.

We distinguish the \emph{prefill} and \emph{decode} phases because the state immediately before answer generation and the states observed during autoregressive decoding follow different distributions. We therefore fit separate operators for the two phases. Given an aligned multimodal conversation, we teacher-force the frozen VLM and extract one representation for each phase, language, sample, and layer.

\paragraph{Prefill representation.}
For each aligned sample, we use the hidden state at the final prompt
position, as the
prefill representation at each layer.

\paragraph{Decode representation.}
Let $\mathcal{T}_{i,\ell}$ contain the teacher-forced answer positions after removing padding, template, and special tokens. We average token-level hidden states into one decoding representation per aligned sample ($\mathbf{h}_{i,\ell,\mathrm{dec}}^{(k)}$), ensuring that longer responses do not contribute more vectors to subspace estimation later.

\paragraph{Fitting corpus.}
We use 4,146 samples from the LLaVA pretraining corpus~\citep{liu2023visual}, reusing existing pretraining data rather than introducing new image-text content.
We translate the English text into ten target languages using GPT-5.5:
Japanese, Korean, Vietnamese, Malay, French, Thai, Hindi, Italian,
Portuguese, and Spanish. This yields strictly aligned multilingual
image--text--answer instances across eleven language views, with English
as the anchor. We use the same samples across models but estimate representations and subspace operators separately for each model.
None of the fitting images or texts appear in the evaluation sets. All operators below are fitted separately for each phase and layer, and we omit $p$ and $k$ when they are clear from context.

\subsection{Multilingual Shared Subspace}
\label{sec:method_gcca}

We next estimate cross-lingually shared structure in centered representations using MAX-VAR generalized canonical correlation analysis (GCCA)~\citep{kettenring1971canonical}. The procedure is performed independently for each phase $p$ and layer $k$.

\paragraph{Language centering and dimensionality reduction.}
For language $\ell$, we stack the representations of its $N$ aligned samples into $\mathbf{X}_\ell \in \mathbb{R}^{N \times d}$ and subtract the language-specific centroid $\boldsymbol{\mu}_\ell$ to obtain the centered matrix $\widetilde{\mathbf{X}}_\ell$.
This allows the subsequent decomposition to operate on instance-dependent variation while preserving the language-control offset during correction.
To identify shared semantic structure across languages, we use GCCA to align their centered representations into a common latent space. To reduce the computational cost of GCCA, we first apply randomized PCA to
each centered matrix, retaining at most $q=64$ principal components, and yielding $\mathbf{Z}_{\ell}\in\mathbb{R}^{N\times q}$ with the projection matrix $\mathbf{A}_{\ell}\in\mathbb{R}^{d\times q}$.

\paragraph{MAX-VAR common latent space.}
We use MAX-VAR GCCA to identify semantic variation consistently expressed across languages, while allowing each language to encode it along different hidden-state directions.
Let $\mathbf{Q}_{\ell}$ be an orthonormal basis for the column space of $\mathbf{Z}_{\ell}$.
GCCA extracts common sample-level structure by taking the leading left singular vectors of the concatenated language-specific bases:
{\setlength{\abovedisplayskip}{5pt}
\setlength{\belowdisplayskip}{5pt}
\setlength{\abovedisplayshortskip}{2pt}
\setlength{\belowdisplayshortskip}{2pt}
\begin{equation}
    \mathbf{G}
    =
    \operatorname{leftSVD}_{r}
    \left(
        [\,\mathbf{Q}_{1}\;\cdots\;\mathbf{Q}_{L}\,]
    \right),
\end{equation}
}
where $L=|\mathcal{L}|$ is the number of language views and $r$ is the shared-subspace rank selected as described in Appendix~\ref{app:rank}.
Importantly, $\mathbf{G}$ is shared in \emph{sample space}. It does not require all languages to use an identical projection in hidden space. For each language, we recover the map from its reduced representation to the common latent variable:
{\setlength{\abovedisplayskip}{5pt}
\setlength{\belowdisplayskip}{5pt}
\setlength{\abovedisplayshortskip}{2pt}
\setlength{\belowdisplayshortskip}{2pt}
\begin{equation}
    \mathbf{W}_{\ell}
    =
    \arg\min_{\mathbf{W}}
    \left\|
        \mathbf{Z}_{\ell}\mathbf{W}
        -
        \mathbf{G}
    \right\|_{F}^{2}.
    \label{eq:wl}
\end{equation}
}
Mapping this solution back to the original hidden space and orthonormalizing yields a language-conditioned shared basis:
{\setlength{\abovedisplayskip}{5pt}
\setlength{\belowdisplayskip}{5pt}
\setlength{\abovedisplayshortskip}{2pt}
\setlength{\belowdisplayshortskip}{2pt}
\begin{equation}
    \mathbf{B}_{\ell}
    =
    \operatorname{orth}
    \left(
        \mathbf{A}_{\ell}\mathbf{W}_{\ell}
    \right)^{\top}
    \in\mathbb{R}^{r\times d}.
\end{equation}
}
We denote the orthogonal projector onto this shared subspace by $\mathbf{P}_{s,\ell}$.
Thus, GCCA aligns languages through common sample-level structure while retaining language-conditioned directions in hidden space.
We select the shared rank $r$ for each phase and layer with a permutation test on a held-out split of the aligned corpus, retaining the leading GCCA dimensions whose weakest cross-language correlation exceeds a permutation-based null threshold. After selection, $\boldsymbol{\mu}_\ell$ and $\mathbf{B}_\ell$ are refit on all 4,146 samples. No downstream hallucination or generation-quality metric is used (Appendix~\ref{app:rank}).

\subsection{Visually Sensitive Subspace}
\label{sec:method_visual}

The estimated multilingual shared subspace captures structure shared across languages, but its limited rank may leave useful visual information outside it. %
We therefore identify visually sensitive directions in its orthogonal complement to preserve during residual attenuation. These directions are estimated from hidden-state changes under image replacement while holding the textual input fixed.

\paragraph{Counterfactual visual differences.}
We select $512$ samples from the fitting corpus.
We compare forward passes using the matched image $I_i$ and a cyclically mismatched image $I_{\pi(i)}$, where $\pi(i)\neq i$, while fixing the text, conversation template, answer tokens, and image-token layout. For each language, phase, and layer, we compute:
{\setlength{\abovedisplayskip}{5pt}
\setlength{\belowdisplayskip}{5pt}
\setlength{\abovedisplayshortskip}{2pt}
\setlength{\belowdisplayshortskip}{2pt}
\begin{equation}
    \mathbf{d}_{i,\ell}
    =
    \mathbf{h}_{i,\ell}
    (I_i,T_{i,\ell})
    -
    \mathbf{h}_{i,\ell}
    (I_{\pi(i)},T_{i,\ell}),
\end{equation}
}
where $T_{i,\ell}$ denotes the textual input of sample $i$ in language $\ell$. These differences capture image-induced hidden-state variation without confounding changes in text or image-token count and position.

\paragraph{Raw visual basis.}
 For each phase and layer, we apply uncentered randomized PCA to the pooled counterfactual differences across fitting samples and language views, obtaining a low-rank basis $\mathbf{U}$ of the dominant visually sensitive directions. The visual rank is selected using held-out differences, as detailed in Appendix~\ref{app:visual_rank}.

\subsection{Strict Visual Protection}
\label{sec:method_strict}

The raw visual basis may overlap with the multilingual shared subspace, so protecting it directly could perturb shared components (see Appendix~\ref{app:seq_proj}). We therefore remove its shared component:
{\setlength{\abovedisplayskip}{5pt}
\setlength{\belowdisplayskip}{5pt}
\setlength{\abovedisplayshortskip}{2pt}
\setlength{\belowdisplayshortskip}{2pt}
\begin{equation}
    \widetilde{\mathbf{U}}_{\ell}
    =
    \mathbf{U}
    \left(
    \mathbf{I}
    -
    \mathbf{P}_{s,\ell}
    \right),
\end{equation}
}
and re-orthonormalize the residual directions, where $\mathbf{I}$ is the $d \times d$ identity matrix.
We denote the projector onto this orthogonalized visual subspace as $\mathbf{P}_{v}$. By construction, visual protection is thus confined to the orthogonal complement of the shared subspace.

\paragraph{Operational decomposition.}
At inference time, we center each hidden state $\mathbf{h}$ by
subtracting the corresponding language-, phase-, and layer-specific centroid, yielding $\boldsymbol{\delta}$. We split $\boldsymbol{\delta}$ as follows, with $\mathbf{I}$ denoting the identity matrix:
{
\setlength{\abovedisplayskip}{5pt}
\setlength{\belowdisplayskip}{5pt}
\setlength{\abovedisplayshortskip}{2pt}
\setlength{\belowdisplayshortskip}{2pt}
\setlength{\jot}{0pt}
\begin{align}
    \mathbf{s} &= \mathbf{P}_{s}\boldsymbol{\delta},
    &\mathbf{c} &= (\mathbf{I}-\mathbf{P}_{s})\boldsymbol{\delta},
    \\
    \mathbf{v} &= \mathbf{P}_{v}\mathbf{c},
    &\mathbf{n} &= (\mathbf{I}-\mathbf{P}_{v})\mathbf{c}.
\end{align}
}
Because $\mathbf{P}_{s}$ and $\mathbf{P}_{v}$ project onto orthogonal subspaces, this yields the exact decomposition
{\setlength{\abovedisplayskip}{5pt}
\setlength{\belowdisplayskip}{5pt}
\setlength{\abovedisplayshortskip}{2pt}
\setlength{\belowdisplayshortskip}{2pt}
\begin{equation}
    \mathbf{h}
    =
    \boldsymbol{\mu}_{\ell}
    +
    \mathbf{s}
    +
    \mathbf{v}
    +
    \mathbf{n}.
\end{equation}
}
Here, $\boldsymbol{\mu}_{\ell}$ is the language offset, $\mathbf{s}$ captures shared semantics, $\mathbf{v}$ the protected visual component, and $\mathbf{n}$ the remaining non-shared residual.
We define the visually protected steering residual as
{\setlength{\abovedisplayskip}{5pt}
\setlength{\belowdisplayskip}{5pt}
\setlength{\abovedisplayshortskip}{2pt}
\setlength{\belowdisplayshortskip}{2pt}
\begin{equation}
    \mathbf{r}_{\gamma}
    =
    \mathbf{c}
    -
    \gamma\mathbf{P}_{v}\mathbf{c}
    =
    (1-\gamma)\mathbf{v}
    +
    \mathbf{n},
    \label{equ:r}
\end{equation}
}
where $\gamma\in[0,1]$ controls the degree of visual protection. When $\gamma=0$, the intervention acts on the entire shared-space complement; when $\gamma=1$, the identified visual component is fully excluded and only $\mathbf{n}$ remains. We use $\gamma=0.5$ throughout our experiments.

\subsection{Norm-Scaled Autoregressive Intervention}
\label{sec:method_inference}

The butterfly-effect hypothesis predicts that individually small deviations can accumulate through autoregressive feedback. We therefore apply the intervention not as a single large correction, but repeatedly to the final prefill state and to subsequent answer-generation states.

A fixed steering coefficient is unsuitable because both
$\|\mathbf{h}\|_2$
and
$\|\mathbf{r}_{\gamma}\|_2$
can vary substantially across languages, layers, phases, and model architectures. We instead control the update through a target relative perturbation ratio $\rho_{\ell}$.
For each intervened state, we compute
{\setlength{\abovedisplayskip}{5pt}
\setlength{\belowdisplayskip}{5pt}
\setlength{\abovedisplayshortskip}{2pt}
\setlength{\belowdisplayshortskip}{2pt}
\begin{equation}
    a_{\ell,p,t}^{(k)}
    =
    \min
    \left(
        a_{\max},
        \rho_{\ell}w_k
        \frac{
            \|\mathbf{h}_{\ell,p,t}^{(k)}\|_2
        }{
            \|\mathbf{r}_{\gamma,\ell,p,t}^{(k)}\|_2
            +
            \epsilon
        }
    \right),
\end{equation}
}
and update the residual-stream state as
\begin{equation}
    \mathbf{h}_{\ell,p,t}^{\prime(k)}
    =
    \mathbf{h}_{\ell,p,t}^{(k)}
    -
    a_{\ell,p,t}^{(k)}
    \mathbf{r}_{\gamma,\ell,p,t}^{(k)}.
\end{equation}

We set $a_{\max}=0.5$ to prevent unusually small residual norms from producing excessively large coefficients. Without clipping, the norm of the update is exactly $\rho_{\ell}w_k$ times the norm of the current state, so the intervention strength is always set relative to the scale of that state. We use RMS-normalized depth-dependent weights $w_k$ to emphasize central intervention layers while maintaining comparable perturbation scales across architectures, as detailed in Appendix~\ref{app:layer_weight}.

The multilingual and visual subspaces, their ranks, and all associated projection operators are estimated without downstream task labels. The language-specific perturbation ratio $\rho_{\ell}$ is selected on a separate calibration set and fixed before evaluation, as described in Appendix~\ref{app:rho_calibration}. At inference time, MVP-Steer therefore requires no model training or parameter update and modifies only the residual-stream states used during answer generation.

\section{Experimental Results}

\paragraph{Experimental setup.}
We evaluate MVP-Steer on Gemma-3-12B-IT, InternVL3.5-14B-IT, and Qwen3-VL-8B-IT with queries in the ten target languages on three benchmarks. AMBER measures object coverage and hallucination in image descriptions, LLaVA-Bench measures general generation quality on open questions about images, and Object HalBench measures object hallucination and object recall in detailed descriptions (metric details in App.~\ref{app:evaluation_metrics}). We compare MVP-Steer with VCD, OPERA, AdaIAT, and CLAIM on InternVL3.5-14B-IT, and pair each method with a baseline run under the same decoding configuration.

\paragraph{Comparison with base multilingual VLMs.}
Table~\ref{tab:main_results} reports macro-averages across the ten target languages. MVP-Steer reduces AMBER CHAIR by 8.84\%, 7.00\%, and 13.77\% on Gemma-3-12B-IT, Qwen3-VL-8B-IT, and InternVL3.5-14B-IT, respectively. Hal and Cog decrease by 6.00\%--12.60\% and 6.94\%--11.79\%, respectively, with all AMBER hallucination reduction intervals excluding zero. On Object HalBench, both CHAIR$_s$ and CHAIR$_i$ decrease across all three backbones, with intervals excluding zero for Gemma-3-12B-IT and InternVL3.5-14B-IT. These reductions preserve object coverage: AMBER Cover changes by less than 0.2\%, while Object HalBench Recall increases by 0.87\%--1.77\%, with all recall improvement intervals excluding zero. Thus, reduced hallucination does not come at the cost of mentioning fewer objects. LLaVA-Bench Overall remains comparable to baseline, with slightly higher point estimates across all three backbones.

\paragraph{Comparison with existing hallucination mitigation methods.}
Compared with existing training-free methods, MVP-Steer shows a more favorable balance between hallucination reduction and generation quality. VCD slightly improves object recall but worsens most hallucination metrics and overall generation quality in our multilingual setting. AdaIAT and CLAIM reduce hallucination on InternVL3.5-14B-IT, but their gains are accompanied by noticeable drops in LLaVA-Bench performance and object recall, while OPERA does not yield consistent improvements in our multilingual setting.
In contrast, MVP-Steer reduces hallucination without sacrificing these broader quality and grounding measures.
Language-level results further show improvements across most language--metric combinations on AMBER and Object HalBench. Detailed per-language comparisons are provided in App.~\ref{app:Per-Language Results of MVP-Steer}.

These results support our central hypothesis that small, repeated corrections to language-dependent representational components can reduce hallucination accumulated during autoregressive generation
while preserving the semantic and visual information required for grounded generation.

\paragraph{Per-language results.}
AMBER CHAIR decreases in all 30 model--language combinations, indicating consistent reductions across languages and backbones. Object HalBench results and AMBER Cover vary more by language: for example, CHAIR$_s$ increases for Korean on InternVL3.5-14B-IT and for Japanese, Spanish, and Vietnamese on Qwen3-VL-8B-IT. App.~\ref{app:Per-Language Results of MVP-Steer} reports baseline comparisons for each backbone.

\begin{figure}[htbp]
    \centering
    \captionsetup{skip=2pt}
    \includegraphics[width=\linewidth]
    {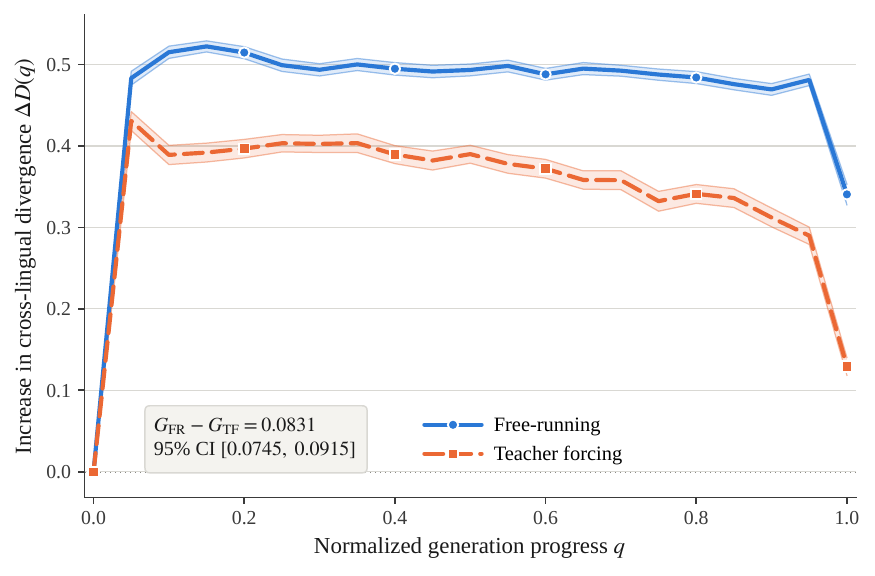}
    \caption{
    Cross-lingual trajectory divergence, $\Delta D(q)=D(q)-D(0)$, under free-running and teacher-forced decoding over normalized generation progress. Shading denotes 95\% confidence intervals.
    }
    \label{fig:trajectory_divergence}
\end{figure}

\begin{table}[htbp]
\centering
\scriptsize
\setlength{\tabcolsep}{0.9pt}
\renewcommand{\arraystretch}{1.08}

\newcommand{\valpct}[2]{%
    \shortstack[c]{#1\\[-1pt]{\tiny $(#2)$}}%
}

\begin{tabular}{@{}lcccccc@{}}
\toprule
Condition &
$E/E_{\mathrm{ref}}$ &
Clip &
Cover $\uparrow$ &
CHAIR $\downarrow$ &
Hal $\downarrow$ &
Cog $\downarrow$ \\
\midrule

Baseline
& --
& --
& 48.60
& 4.43
& 12.94
& 2.83 \\

\addlinespace[2pt]

Repeated
& 1.0000
& 0\%
& \valpct{48.46}{-0.08\%}
& \valpct{3.85}{+13.68\%}
& \valpct{11.36}{+12.55\%}
& \valpct{2.52}{+10.79\%} \\

\addlinespace[2pt]

Initial only
& 1.0045
& 74.75\%
& \valpct{47.77}{-1.59\%}
& \valpct{3.83}{+14.10\%}
& \valpct{10.96}{+13.46\%}
& \valpct{2.42}{+14.37\%} \\

\addlinespace[2pt]

Every 4
& 1.0071
& 0\%
& \valpct{48.73}{+0.48\%}
& \valpct{4.13}{+6.11\%}
& \valpct{12.08}{+4.92\%}
& \valpct{2.64}{+4.61\%} \\

\bottomrule
\end{tabular}
\caption{
Budget-matched temporal intervention on 804 evaluation images.
$E/E_{\mathrm{ref}}$ denotes relative perturbation energy, and Clip denotes the percentage of intervened states whose update is clipped by $a_{\max}$.
Values in parentheses show relative changes from the baseline.
}
\label{tab:budget_matched_temporal}
\end{table}

\section{Butterfly Effect and Temporal Intervention Analysis}

\paragraph{Cross-lingual trajectory divergence.}
To visualize autoregressive amplification, we compare free-running and teacher-forced hidden-state trajectories using identical images and equivalent multilingual prompts. Free-running generation produces an early separation that remains elevated, while teacher-forced trajectories progressively reconverge, as shown in Figure~\ref{fig:trajectory_divergence}. The difference in early-to-late divergence change is $0.0831$, with the same overall direction across all ten target languages. Appendix~\ref{app:butterfly_empirical} describes the trajectory construction and statistical analysis.

\paragraph{From local corrections to sequence-level effects.} MVP-Steer repeatedly applies small representation-level corrections across decoding steps and intervention layers, with each unclipped update having norm $\rho_\ell w_k$ times the hidden-state norm. We quantify intervention energy as the sum of squared update norms across steps and layers. Since earlier corrections propagate through subsequent computation, their temporal distribution can affect sequence-level outcomes beyond total energy (Appendix~\ref{app:butterfly_derivation}). We test this on InternVL3.5-14B-IT using 804 AMBER images, comparing token-wise correction with two schedules whose energies match within 1\%: correcting only the first decoding step or every fourth token.

As shown in Table~\ref{tab:budget_matched_temporal}, token-wise correction reduces CHAIR by 13.68\%, more than twice the reduction achieved by every-four-token correction. Initial-only correction achieves comparable hallucination reduction but incurs substantial clipping and reduced object coverage. These findings indicate that distributing small corrections throughout decoding provides a better balance between hallucination reduction and visual coverage, consistent with cumulative effects during autoregressive generation. Appendix~\ref{app:butterfly_derivation} provides the propagation analysis, with additional experimental details in Appendix~\ref{app:butterfly_empirical}. Under the same per-state budget, random, shared-only and visual-only directions reduce CHAIR by at most 2.7\%, compared with 13.8\% for MVP-Steer
(Appendix~\ref{app:Effect of the Steering Direction}).

\paragraph{Layer-wise alignment analysis.}
We measure cross-lingual alignment on InternVL3.5-14B-IT using
teacher-forced final-answer-token representations and
final-prompt-token visual attention. Attention is taken from the
Top-5 visually sensitive heads selected on an independent
calibration set. Across L14--27, the alignment gap drops on average by $1.82\%$
for representations and $1.91\%$ for attention, with gains at
all 14 layers (metric, full results, and details in
Appendix~\ref{app:Detailed layer-wise cross-lingual alignment analysis}).
Free-running divergence also exceeds teacher-forced divergence, and
schedules with similar mean energy yield different
hallucination--coverage trade-offs. These findings, together with
the modest but consistent gains, are in line with autoregressive
amplification and motivate repeated correction during generation.

\section{Conclusion}

Our study provides a representation-level account of multilingual visual hallucination, showing that cross-lingual disparities may emerge from small but systematic language-dependent deviations that compound throughout autoregressive generation. Motivated by this finding, we propose MVP-Steer, a training-free inference-time method that selectively attenuates non-shared residuals while preserving language anchors, shared semantics, and visually grounded information. Across three VLMs, ten languages, and three benchmarks, MVP-Steer consistently reduces hallucination without compromising generation quality or visual grounding, demonstrating the potential of targeted representation steering for improving multilingual VLM reliability.

\section*{Limitations}

\paragraph{Estimation from limited data.}
The language centroids, the shared and visually sensitive subspaces, and their ranks are estimated from 4,146 aligned samples and 512 counterfactual image pairs, and the target-language text is produced by machine translation. These finite samples cannot recover the underlying representation structure exactly, so the decomposition into shared, visual, and residual components contains estimation error. Some shared semantic or visual information may remain in the attenuated residual, and some language-dependent variation may fall outside it. Larger and more diverse fitting data, such as human-written multilingual descriptions, may yield more accurate subspace estimates.

\paragraph{Access and calibration requirements.}
MVP-Steer reads and modifies residual-stream states, so it applies only to open-weight models and cannot be used with VLMs served through closed APIs. It also requires fitting operators and calibrating $\rho_\ell$ for each backbone and language, which adds setup cost when extending to new models or languages. We evaluate three backbones with 8B to 14B parameters, and its behavior on larger models or other architectures remains untested.

\paragraph{Scope of evaluation and analysis.}
Our evaluation covers ten languages and mainly measures object hallucination, using queries derived from English benchmarks. Lower-resource languages and attribute- or relation-level hallucination are not covered. Due to computational resource constraints, we focus most mechanistic analyses, including trajectory and temporal-intervention experiments, on InternVL3.5-14B-IT. Extending these analyses to additional backbones and larger multilingual settings remains future work. These results are consistent with the butterfly-effect hypothesis but do not show that it is the only source of cross-lingual hallucination disparities.

 \section*{Ethical Statement}
\label{sec:ethical-statement}

Our study investigates cross-lingual disparities in visual hallucination across ten languages using existing image–text datasets and benchmarks. These disparities reflect model behavior under specific evaluation conditions and should not be interpreted as differences in the expressive capabilities or reliability of individual languages. The multilingual fitting corpus is constructed by translating English descriptions into ten target languages using GPT-5.5. While these translations generally preserve the intended semantics, they may not fully capture the linguistic nuances and natural expressions of each target language. Consequently, minor differences in expression may remain across otherwise semantically aligned multilingual samples.

MVP-Steer modifies intermediate hidden representations during inference to reduce language-dependent residual variation. Although the method preserves language centroids, shared semantic components, and visually sensitive directions, imperfect subspace estimation may inadvertently suppress information needed for accurate descriptions or language-specific expression. Such interventions could introduce unintended changes in generated content, particularly for languages, visual domains, or model architectures beyond those evaluated. We constrain intervention strength through language-specific calibration and evaluate both hallucination reduction and generation quality, while recognizing that these safeguards do not guarantee reliable behavior in all deployment settings.

Our evaluation relies on established benchmarks and automated metrics that primarily assess object hallucination and general response quality. These measures may overlook subtle semantic errors, culturally dependent interpretations, or other forms of visual misinformation. GPT-5.5 is also used for multilingual object extraction during calibration, which may introduce additional language-dependent measurement errors. The datasets, benchmarks, and pretrained models used in this study remain subject to their respective licensing and usage conditions. Although MVP-Steer improves visual grounding across the evaluated settings, its results should not be considered a guarantee of factual correctness or sufficient evidence for deployment in safety-critical applications.

AI assistants supported manuscript editing and polishing, with all scientific content reviewed and verified by the authors.

\bibliography{custom}

\appendix

\section{Detailed Representation Geometry Analysis}
\label{app:geometry_analysis}

This section provides the complete experimental definitions and results underlying Observations~1--2 in Section~\ref{sec:Motivating Observations}.

\paragraph{Representation construction.}
We use 4,146 aligned image--description instances in ten target languages together with English. For semantic instance $i$, language $\ell$, and transformer layer $k$, we represent the description by mean-pooling the hidden states associated with its answer tokens, excluding special and template tokens:
\[
\mathbf{h}_{i,\ell}^{(k)}
=
\frac{1}{|\mathcal{T}_{i,\ell}|}
\sum_{t\in\mathcal{T}_{i,\ell}}
\mathbf{H}_{i,\ell,t}^{(k)}.
\]
InternVL3.5-14B-IT is analyzed over its primary layer range L14--27, while Qwen3-VL-8B-IT is analyzed over L13--24. Unless otherwise stated, reported statistics are averaged over the corresponding primary layer range.

\paragraph{Relational semantic geometry.}
Observation~1 evaluates whether different languages preserve the same relational organization among semantic instances rather than only point-wise correspondence. For each language $\ell$, we construct a within-language representational similarity matrix
\[
\mathbf{R}_{\ell}(i,j)
=
\cos\left(
\mathbf{h}_{i,\ell},
\mathbf{h}_{j,\ell}
\right).
\]
For a pair of languages $(\ell,m)$, representational similarity analysis (RSA)~\citep{kriegeskorte2008representational} compares the upper-triangular entries of their similarity matrices:
\[
\mathrm{RSA}(\ell,m)
=
\operatorname{Spearman}
\left(
\operatorname{vec}_{\triangle}(\mathbf{R}_{\ell}),
\operatorname{vec}_{\triangle}(\mathbf{R}_{m})
\right).
\]
We compute RSA for all English-to-target and target-to-target language pairs and average the result over the analyzed layers.

\begin{table}[t]
\centering
\small

\setlength{\tabcolsep}{5.5pt}
\renewcommand{\arraystretch}{1.08}

\begin{tabular}{@{}lcc@{}}
\toprule
\textbf{Language}
&
\textbf{InternVL3.5-14B-IT}
&
\textbf{Qwen3-VL-8B-IT}
\\
\midrule

French
& 0.7844 (0.0545)
& 0.8347 (0.0238)
\\

Hindi
& 0.6453 (0.0670)
& 0.6912 (0.0411)
\\

Italian
& 0.7754 (0.0597)
& 0.8285 (0.0240)
\\

Japanese
& 0.6794 (0.0776)
& 0.7446 (0.0332)
\\

Korean
& 0.6803 (0.0795)
& 0.7429 (0.0320)
\\

Malay
& 0.6791 (0.0740)
& 0.7677 (0.0246)
\\

Portuguese
& 0.7898 (0.0558)
& 0.8376 (0.0223)
\\

Spanish
& 0.7885 (0.0574)
& 0.8418 (0.0223)
\\

Thai
& 0.6331 (0.0683)
& 0.6931 (0.0303)
\\

Vietnamese
& 0.7265 (0.0651)
& 0.7856 (0.0263)
\\

\midrule
\textbf{Average}
& \textbf{0.7182}
& \textbf{0.7768}
\\

\bottomrule
\end{tabular}
\caption{
English-to-target representational similarity analysis (RSA).
Values report the mean RSA across the analyzed layers, with the cross-layer
standard deviation shown in parentheses.
InternVL3.5-14B-IT is analyzed over Layers 14 through 27, while
Qwen3-VL-8B-IT is analyzed over Layers 13 through 24.
}
\label{tab:rsa_languagewise}
\end{table}

\begin{table}[t]
\centering
\small

\setlength{\tabcolsep}{5pt}
\renewcommand{\arraystretch}{1.08}

\begin{tabular}{@{}lccc@{}}
\toprule

\textbf{Model}
&
\textbf{En--Tgt. (10)}
&
\textbf{Tgt.--Tgt. (45)}
&
\textbf{All (55)}
\\

\midrule

InternVL3.5-14B-IT
& 0.7182
& \textbf{0.7615}
& 0.7536
\\

Qwen3-VL-8B-IT
& 0.7768
& \textbf{0.7740}
& 0.7745
\\

\bottomrule
\end{tabular}
\caption{
RSA across different cross-lingual language-pair sets.
}
\label{tab:rsa_pair_summary}
\end{table}

\begin{table*}[t]
\centering
\small

\setlength{\tabcolsep}{3.3pt}
\renewcommand{\arraystretch}{1.08}

\begin{tabular}{@{}lcccccccc@{}}
\toprule

\multirow{2}{*}{\textbf{Language}}
&
\multicolumn{4}{c}{\textbf{Raw Representation}}
&
\multicolumn{4}{c}{\textbf{Centroid Removed}}
\\

\cmidrule(lr){2-5}
\cmidrule(lr){6-9}

&
\textbf{Match}
&
\textbf{Mismatch}
&
\textbf{Margin}
&
\textbf{R@1}
&
\textbf{Match}
&
\textbf{Mismatch}
&
\textbf{Margin}
&
\textbf{R@1}
\\

\midrule

French
& 0.8360 & 0.5910 & 0.2451 & 0.9952
& 0.7153 & 0.0005 & 0.7148 & 0.9999
\\

Hindi
& 0.6796 & 0.5361 & 0.1435 & 0.8850
& 0.5576 & 0.0006 & 0.5571 & 0.9997
\\

Italian
& 0.8255 & 0.5866 & 0.2389 & 0.9928
& 0.7059 & 0.0006 & 0.7053 & 0.9998
\\

Japanese
& 0.7389 & 0.5302 & 0.2087 & 0.9810
& 0.6181 & 0.0004 & 0.6176 & 0.9998
\\

Korean
& 0.7425 & 0.5481 & 0.1944 & 0.9735
& 0.6074 & 0.0005 & 0.6069 & 0.9997
\\

Malay
& 0.7755 & 0.5794 & 0.1961 & 0.9741
& 0.6190 & 0.0006 & 0.6184 & 0.9995
\\

Portuguese
& 0.8293 & 0.5875 & 0.2418 & 0.9957
& 0.7132 & 0.0006 & 0.7126 & 1.0000
\\

Spanish
& 0.8348 & 0.5883 & 0.2465 & 0.9953
& 0.7200 & 0.0005 & 0.7195 & 1.0000
\\

Thai
& 0.6971 & 0.5248 & 0.1723 & 0.9322
& 0.5733 & 0.0004 & 0.5728 & 0.9994
\\

Vietnamese
& 0.7866 & 0.5757 & 0.2109 & 0.9846
& 0.6453 & 0.0005 & 0.6447 & 0.9997
\\

\midrule

\textbf{Average}
& \textbf{0.7746}
& \textbf{0.5648}
& \textbf{0.2098}
& \textbf{0.9709}
& \textbf{0.6475}
& \textbf{0.0005}
& \textbf{0.6470}
& \textbf{0.9998}
\\

\bottomrule
\end{tabular}
\caption{
Language-wise representation geometry on InternVL3.5-14B-IT before and after
language centroid removal.
Match and Mismatch denote cosine similarity between semantically matched and
mismatched cross-lingual instances, respectively.
Margin is defined as Match minus Mismatch.
R@1 denotes bidirectional cross-lingual retrieval Recall@1.
}
\label{tab:centering_internvl}
\end{table*}

\begin{table*}[t]
\centering
\small

\setlength{\tabcolsep}{3.3pt}
\renewcommand{\arraystretch}{1.08}

\begin{tabular}{@{}lcccccccc@{}}
\toprule

\multirow{2}{*}{\textbf{Language}}
&
\multicolumn{4}{c}{\textbf{Raw Representation}}
&
\multicolumn{4}{c}{\textbf{Centroid Removed}}
\\

\cmidrule(lr){2-5}
\cmidrule(lr){6-9}

&
\textbf{Match}
&
\textbf{Mismatch}
&
\textbf{Margin}
&
\textbf{R@1}
&
\textbf{Match}
&
\textbf{Mismatch}
&
\textbf{Margin}
&
\textbf{R@1}
\\

\midrule

French
& 0.9311 & 0.8158 & 0.1153 & 0.9966
& 0.7534 & 0.0009 & 0.7525 & 0.9998
\\

Hindi
& 0.8311 & 0.7563 & 0.0748 & 0.8859
& 0.5962 & 0.0010 & 0.5952 & 0.9997
\\

Italian
& 0.9272 & 0.8137 & 0.1135 & 0.9965
& 0.7453 & 0.0010 & 0.7443 & 1.0000
\\

Japanese
& 0.8780 & 0.7739 & 0.1040 & 0.9719
& 0.6630 & 0.0008 & 0.6623 & 0.9996
\\

Korean
& 0.8648 & 0.7677 & 0.0971 & 0.9526
& 0.6371 & 0.0009 & 0.6362 & 0.9999
\\

Malay
& 0.9032 & 0.8044 & 0.0987 & 0.9795
& 0.6759 & 0.0010 & 0.6748 & 0.9998
\\

Portuguese
& 0.9301 & 0.8157 & 0.1144 & 0.9966
& 0.7547 & 0.0010 & 0.7537 & 1.0000
\\

Spanish
& 0.9324 & 0.8168 & 0.1156 & 0.9978
& 0.7607 & 0.0010 & 0.7598 & 1.0000
\\

Thai
& 0.8467 & 0.7582 & 0.0884 & 0.8810
& 0.6126 & 0.0008 & 0.6118 & 0.9994
\\

Vietnamese
& 0.9037 & 0.8029 & 0.1008 & 0.9795
& 0.6861 & 0.0009 & 0.6852 & 0.9990
\\

\midrule

\textbf{Average}
& \textbf{0.8948}
& \textbf{0.7926}
& \textbf{0.1023}
& \textbf{0.9638}
& \textbf{0.6885}
& \textbf{0.0009}
& \textbf{0.6876}
& \textbf{0.9997}
\\

\bottomrule
\end{tabular}
\caption{
Language-wise representation geometry on Qwen3-VL-8B-IT before and after
language centroid removal.
Match and Mismatch denote cosine similarity between semantically matched and
mismatched cross-lingual instances, respectively.
Margin is defined as Match minus Mismatch.
R@1 denotes bidirectional cross-lingual retrieval Recall@1.
}
\label{tab:centering_qwen}
\end{table*}

Across the 45 target-to-target pairs, mean RSA reaches $0.7615$ for InternVL3.5-14B-IT and $0.7740$ for Qwen3-VL-8B-IT. When English is included, the mean over all 55 language pairs is $0.7536$ for InternVL and $0.7745$ for Qwen. English-to-target averages are $0.7182$ and $0.7768$, respectively. Tables~\ref{tab:rsa_languagewise} and
\ref{tab:rsa_pair_summary} report the complete language-wise and pair-range results, including cross-layer variation.

\paragraph{Matched and mismatched semantic similarity.}
We additionally evaluate point-wise semantic correspondence. For a language pair $(\ell,m)$, matched similarity is defined as
\[
s_{\ell,m}^{+}
=
\frac{1}{N}
\sum_{i=1}^{N}
\cos\left(
\mathbf{h}_{i,\ell},
\mathbf{h}_{i,m}
\right),
\]
whereas mismatched similarity measures the background similarity between different semantic instances:
\[
s_{\ell,m}^{-}
=
\frac{1}{N(N-1)}
\sum_{i\neq j}
\cos\left(
\mathbf{h}_{i,\ell},
\mathbf{h}_{j,m}
\right).
\]
We define the semantic margin as
\[
M_{\ell,m}
=
s_{\ell,m}^{+}
-
s_{\ell,m}^{-}.
\]
Bidirectional Recall@1 is computed by retrieving the corresponding semantic instance across the two languages in both directions.

Before centering, English-to-target Recall@1 averages $0.9709$ on InternVL and $0.9638$ on Qwen, showing that semantic counterparts are already reliably identifiable in the raw representation space.

\paragraph{Language-centroid removal.}
To isolate the effect of the language-level offset, we estimate a centroid independently for each language and layer:
\[
\boldsymbol{\mu}_{\ell}^{(k)}
=
\frac{1}{N_{\mathrm{fit}}}
\sum_{i=1}^{N_{\mathrm{fit}}}
\mathbf{h}_{i,\ell}^{(k)}.
\]
The centered representation is
\[
\widetilde{\mathbf{h}}_{i,\ell}^{(k)}
=
\mathbf{h}_{i,\ell}^{(k)}
-
\boldsymbol{\mu}_{\ell}^{(k)}.
\]
Centroids are estimated from 2,085 fitting instances, while all reported centering metrics are computed on a disjoint set of 2,061 instances.

For InternVL, the average semantic margin across the 45 target-to-target pairs increases from $0.1806$ in the raw space to $0.6648$ after centroid removal, while mismatched cosine decreases from $0.6451$ to $0.0006$. For Qwen, the corresponding margin increases from $0.0939$ to $0.6876$, while mismatched cosine decreases from $0.8114$ to $0.0010$. English-to-target Recall@1 reaches $0.9998$ for InternVL and $0.9997$ for Qwen after centering.

These results separate two aspects of multilingual representation geometry. Relational semantic organization is already strongly shared across languages, as reflected by high RSA and raw retrieval accuracy, whereas the language centroid shifts this geometry within the ambient representation space. Centroid removal exposes the shared relational structure more directly by suppressing background similarity between semantically unrelated instances.

Complete per-language raw and centered matched similarity, mismatched similarity, semantic margin, and Recall@1 are reported in Tables~\ref{tab:centering_internvl} and \ref{tab:centering_qwen}.

\section{Causal Intervention on Language Centroids}
\label{app:centroid_intervention}

To further test the causal role of language-conditioned centroids, we directly shift hidden representations between each target language and English.
The experiment is conducted on InternVL3.5-14B-IT using Japanese, Korean, and Vietnamese, with 200 AMBER samples per language.

\paragraph{Centroid-shift direction.}
For each target language $\ell$ and layer $k$, we estimate the English and target-language centroids,
$\boldsymbol{\mu}_{\mathrm{en}}^{(k)}$ and
$\boldsymbol{\mu}_{\ell}^{(k)}$,
from the final prompt-token representations of 512 aligned LLaVA~\citep{liu2024improved} examples.
We construct a scale-normalized English-transfer direction as
\begin{equation}
    \mathbf{d}_{\ell}^{(k)}
    =
    \frac{
        \boldsymbol{\mu}_{\mathrm{en}}^{(k)}
        -
        \boldsymbol{\mu}_{\ell}^{(k)}
    }{
        \left\|
        \boldsymbol{\mu}_{\mathrm{en}}^{(k)}
        -
        \boldsymbol{\mu}_{\ell}^{(k)}
        \right\|_2
    }
    \left\|
        \boldsymbol{\mu}_{\ell}^{(k)}
    \right\|_2 .
\end{equation}
During generation, the intervention is applied only to textual prefill and decoding states; image and padding tokens are left unchanged.

\paragraph{Intermediate-layer language transfer.}
We first add
$0.05\mathbf{d}_{\ell}^{(k)}$
throughout layers 14--27.
Although the resulting perturbation is small relative to the hidden-state norm
($\rho \approx 0.0032$--$0.0036$),
the intervention switches all generated responses from Japanese, Korean, and Vietnamese to English.
Importantly, the English outputs retain non-trivial visual-description capability rather than collapsing into invalid text.
For example, after the shift, English-evaluated Cover remains 0.5516, 0.5250, and 0.5485 for Japanese, Korean, and Vietnamese inputs, respectively.
These results show that the centroid difference is not merely correlational: a small displacement along this direction is sufficient to causally redirect the output language.

\begin{table*}[htbp]
\centering
\small

\setlength{\tabcolsep}{5pt}
\renewcommand{\arraystretch}{1.15}

\begin{tabular}{lcccccc}
\toprule
& \multicolumn{2}{c}{\textbf{Original}}
& \multicolumn{4}{c}{\textbf{Mid-layer shift: $+0.05\mathbf{d}_{\ell}$}} \\
\cmidrule(lr){2-3}
\cmidrule(lr){4-7}
\textbf{Language}
& \textbf{Target (\%)}
& \textbf{Target-language metrics}
& $\boldsymbol{\rho}$
& \textbf{Target (\%)}
& \textbf{EN (\%)}
& \textbf{English metrics} \\
\midrule

Japanese
& 100.0
& .5409 / .0413 / .105 / .0223
& .00364
& 0.0
& 100.0
& .5516 / .0631 / .170 / .0373 \\

Korean
& 99.5
& .4393 / .0586 / .160 / .0340
& .00321
& 0.0
& 100.0
& .5250 / .0440 / .115 / .0271 \\

Vietnamese
& 100.0
& .5922 / .0541 / .155 / .0358
& .00317
& 0.0
& 100.0
& .5485 / .0436 / .130 / .0297 \\

\bottomrule
\end{tabular}
\caption{
Effect of intermediate-layer centroid shifting on InternVL3.5-14B-IT.
We add $0.05\mathbf{d}_{\ell}^{(k)}$ to textual hidden states at layers 14--27.
$\rho$ denotes the realized relative perturbation magnitude.
Target (\%) and EN (\%) report the proportions of responses generated in the original target language and English, respectively.
Performance is reported as Cover / CHAIR / Hal / Cog.
After the intervention, metrics are evaluated using the English vocabulary because all responses switch to English.
}
\label{tab:centroid_mid}
\end{table*}

\paragraph{Sensitivity to the reverse-shift strength.}
Because the intermediate and upper-layer interventions operate over different
layer ranges and their centroid-difference directions have different layer-wise
scales, the numerical coefficients of the two interventions are not expected
to cancel one another directly.
We therefore keep the intermediate-layer shift fixed at
$+0.05\mathbf{d}_{\ell}^{(k)}$ for layers 14--27 and sweep the upper-layer
reverse-shift strength
$\beta\in\{0.0125,0.025,0.05,0.10,0.20\}$,
applying $-\beta\mathbf{d}_{\ell}^{(k)}$ at layers 29--39.

As shown in Table~\ref{tab:centroid_strength_sweep}, weak reverse shifts
($\beta\leq0.025$) fail to restore the target language, despite already
producing a cumulative perturbation magnitude comparable to that of the
intermediate intervention.
Target-language realization begins to recover around $\beta=0.05$ and becomes
substantial at $\beta=0.10$.
Increasing the strength further does not recover the original generation
capability: although $\beta=0.20$ yields high target-language rates, descriptive
coverage collapses almost completely.
Across all tested strengths, we therefore observe no setting that simultaneously
restores both the original output language and the baseline visual-description
capability.

\begin{table*}[t]
\centering
\small

\setlength{\tabcolsep}{4.5pt}
\renewcommand{\arraystretch}{1.15}

\begin{tabular}{lcccccc}
\toprule
& \multicolumn{2}{c}{\textbf{Japanese}}
& \multicolumn{2}{c}{\textbf{Korean}}
& \multicolumn{2}{c}{\textbf{Vietnamese}} \\
\cmidrule(lr){2-3}
\cmidrule(lr){4-5}
\cmidrule(lr){6-7}

$\boldsymbol{\beta}$
& Target (\%) & Cover
& Target (\%) & Cover
& Target (\%) & Cover \\
\midrule

Baseline
& 100.0 & .5409
& 99.5 & .4393
& 100.0 & .5922 \\

0.0125
& 0.0 & --
& 0.0 & --
& 0.0 & -- \\

0.025
& 0.0 & --
& 0.0 & --
& 0.0 & -- \\

0.050
& 4.0 & .1140
& 41.5 & .0390
& 9.0 & .0916 \\

0.100
& 70.0 & .0989
& 100.0 & .0486
& 100.0 & .0912 \\

0.200
& 91.0 & .0046
& 100.0 & .0000
& 83.0 & .0000 \\

\bottomrule
\end{tabular}
\caption{
Sensitivity to the upper-layer reverse-shift strength on InternVL3.5-14B-IT.
The intermediate intervention is fixed to
$+0.05\mathbf{d}_{\ell}^{(k)}$ at layers 14--27, followed by
$-\beta\mathbf{d}_{\ell}^{(k)}$ at layers 29--39.
Target (\%) denotes the proportion of responses generated in the original
target language, and Cover is evaluated with the corresponding target-language
vocabulary.
Increasing $\beta$ eventually restores the surface language, but no tested
strength recovers the original descriptive coverage.
}
\label{tab:centroid_strength_sweep}
\end{table*}

\paragraph{Late-layer recovery at a representative strength.}
We next ask whether reversing the centroid displacement at later layers can recover both the original output language and the generation capability altered by the intermediate-layer intervention.
Starting from the English-switched state, we inject the opposite direction,
$-0.10\mathbf{d}_{\ell}^{(k)}$,
throughout layers 29--39.
This late intervention substantially restores the surface language: the target-language rate returns to 70\% for Japanese and 100\% for both Korean and Vietnamese, while the English rate falls to 2.5\%, 0\%, and 0\%, respectively.

However, restoring the output language does not restore the original generation capability.
Target-language Cover drops from 0.5409 to 0.0989 for Japanese, from 0.4393 to 0.0486 for Korean, and from 0.5922 to 0.0912 for Vietnamese.
Moreover, 22\% of Japanese and 54\% of Vietnamese outputs become degenerate under the late reverse intervention.
The very low CHAIR, Hal, and Cog values after recovery therefore reflect a severe loss of descriptive coverage rather than improved grounding.

These results reveal an asymmetry between language realization and semantic generation.
The centroid direction remains causally effective at high layers and can redirect the surface form back to the target language, but it does not recover the semantic and visual generation trajectory altered by the earlier intervention.
This supports the view that language identity can be corrected late, whereas information lost or redirected in intermediate layers is not recoverable through a subsequent centroid shift alone.

\section{Cross-Lingual Visual Grounding Analysis}
\label{app:visual_grounding}

This section provides additional experimental details and results for Observation~3 in Section~\ref{sec:Motivating Observations}.

\begin{table}[htbp]
\centering
\small

\setlength{\tabcolsep}{6pt}
\renewcommand{\arraystretch}{1.10}

\begin{tabular}{@{}lc@{}}
\toprule
\textbf{Metric}
&
\textbf{Similarity}
\\
\midrule

Visual attention cosine
& 0.9514
\\

Jensen--Shannon similarity
& 0.9577
\\

Visual attention mass similarity
& 0.9287
\\

\bottomrule
\end{tabular}
\caption{
Cross-lingual visual grounding consistency on the baseline
InternVL3.5-14B-IT.
Prompt-final metrics are computed from the attention of the final prompt
token to visual tokens over 100 AMBER images and all 45 target language
pairs.
Higher values indicate greater cross-lingual consistency.
}
\label{tab:visual_grounding_summary}
\end{table}

\begin{table}[htbp]
\centering
\small

\setlength{\tabcolsep}{6pt}
\renewcommand{\arraystretch}{1.08}

\begin{tabular}{@{}lcc@{}}
\toprule

\textbf{Language}
&
\textbf{Prompt Final}
&
\textbf{Teacher Forcing}
\\

\midrule

French
& 0.9584
& 0.9660
\\

Hindi
& 0.9414
& 0.9499
\\

Italian
& 0.9584
& 0.9655
\\

Japanese
& 0.9480
& 0.9586
\\

Korean
& 0.9511
& 0.9625
\\

Malay
& 0.9468
& 0.9538
\\

Portuguese
& 0.9566
& 0.9626
\\

Spanish
& 0.9508
& 0.9598
\\

Thai
& 0.9499
& 0.9617
\\

Vietnamese
& 0.9528
& 0.9629
\\

\midrule

\textbf{Average}
& \textbf{0.9514}
& \textbf{0.9603}
\\

\bottomrule
\end{tabular}
\caption{
Language-wise cross-lingual visual attention consistency on the baseline
InternVL3.5-14B-IT.
For each language, similarity is averaged over its pairings with the other
nine target languages.
Prompt Final measures visual attention immediately before generation.
Teacher Forcing measures response-level visual attention under teacher
forcing, aggregated over semantic units within each language.
}
\label{tab:visual_grounding_languagewise}
\end{table}

\paragraph{Prompt-final visual attention.}
We sample 100 images from the held-out portion of AMBER and pair each image with semantically equivalent questions in ten target languages. This produces 45 unordered language pairs for every image. We use exact eager causal self attention and extract, at each analyzed layer, the attention from the final prompt token to all visual tokens. Attention maps are averaged across heads and normalized over the visual token dimension before comparison. Cross-lingual consistency is measured by the cosine similarity between the normalized visual attention maps produced by two languages for the same image.

As shown in Table~\ref{tab:visual_grounding_summary}, the average cross-lingual cosine reaches $0.9514$. Jensen--Shannon similarity is similarly high at $0.9577$, while visual attention mass similarity reaches $0.9287$. These results show that semantically equivalent multilingual queries induce highly similar visual attention patterns before autoregressive generation begins.

\paragraph{Consistency across languages.}
Table~\ref{tab:visual_grounding_languagewise} reports the results for each target language by averaging its similarity with the other nine languages. Prompt-final cosine remains consistently high across all ten languages, ranging from $0.9414$ for Hindi to $0.9584$ for Italian. The high overall similarity therefore reflects a broadly shared visual grounding pattern rather than a small subset of closely related languages.

\paragraph{Teacher-forced response-level analysis.}
We additionally examine whether this consistency persists beyond the initial generation step. We use a canonical description together with semantically equivalent translations and perform teacher forcing throughout the response. Because token positions are not directly comparable across languages with different word orders and tokenizations, we segment each response into semantic units, average visual attention over the tokens within each unit, and then average equally across units to obtain one response-level attention map per language. We then compute the cosine similarity between these maps for each language pair.

The resulting average cross-lingual cosine is $0.9603$. As shown in Table~\ref{tab:visual_grounding_languagewise}, it ranges from $0.9499$ for Hindi to $0.9660$ for French. This result extends the prompt-final observation beyond the initial generation state and shows that visual grounding remains highly consistent when semantic content and generation history are controlled.

\section{Cross-Lingual Visual Attention Under Free-Running Decoding}
\label{app:attention_free_running}

\paragraph{Experimental setup.}
We extend Observation~3 to examine whether cross-lingual
visual attention remains consistent during free-running
generation.
We reuse the same 100 held-out AMBER images and ten target
languages, covering 45 language pairs.
Using InternVL3.5-14B-IT, we analyze visual attention over
layers 13--32 under two decoding conditions:
free running, where each language generates autoregressively
from its own predicted tokens, and teacher forcing, where
reference translated prefixes are provided.

To ensure a comparable analysis despite differences in
generated content, tokenization, and word order, we
independently segment each language's response into semantic
units.
Visual attention is first averaged over tokens within
each unit and then equally averaged across units to obtain
a response-level attention map.
This aggregation, also used for the teacher-forcing result in Observation~3,
does not require cross-lingual alignment
of individual semantic units.
We apply the same procedure to both decoding conditions
and compute pairwise cosine similarity between the resulting
response-level attention maps.

\paragraph{Results and analysis.}
Cross-lingual visual attention remains highly similar
under free-running decoding, achieving an average cosine
similarity of $0.9492$ (95\% CI: $[0.9448, 0.9533]$).
Teacher forcing further increases similarity to $0.9603$,
with a difference of $0.0112$
(95\% CI: $[0.00988, 0.01255]$).
Equivalently, the average cosine distance decreases
from $0.05083$ under free running to $0.03966$
under teacher forcing, corresponding to a $21.97\%$
reduction in cross-lingual attention divergence.
This pattern holds for all 100 images when similarities
are averaged across language pairs at the image level.

The free-running results are also robust to different
aggregation strategies: averaging over all semantic tokens
and all generated tokens yields cosine similarities of
$0.9491$ and $0.9522$, respectively, compared with
$0.9492$ under equal semantic-unit weighting.
The corresponding Jensen--Shannon and visual-mass
similarities under unit-weighted aggregation are
$0.9601$ and $0.9370$, respectively.
Together, these findings indicate that visual attention
remains strongly aligned across languages, while
unconstrained generation is associated with a modest
but consistent increase in cross-lingual variation.

\section{Empirical Validation of Autoregressive Amplification}
\label{app:butterfly_empirical}

To complement the trajectory-level derivation in Appendix~\ref{app:butterfly_derivation}, we examine three aspects of autoregressive amplification. We first analyze how object hallucinations vary across generation positions, then compare free-running and teacher-forced representation trajectories, and finally evaluate the effect of temporal intervention placement under approximately matched perturbation budgets.

\paragraph{Position-wise object hallucination: setup.}
We reuse the same 100 held-out AMBER images from Observation~3 and analyze free-running descriptions generated by InternVL3.5-14B-IT across the ten target languages. Each response is divided into three approximately equal segments by token position, yielding comparable token counts of 42,101, 42,406, and 42,817, respectively. Using the AMBER object vocabulary, we measure hallucination mention rates, hallucinated mentions per 100 tokens, and the fraction of responses containing hallucinated objects. Uncertainty is estimated using paired image-level bootstrap.

\paragraph{Position-wise object hallucination: results.}
Table~\ref{tab:position_hallucination} shows that the language-averaged hallucination mention rate increases from 1.43\% in the first third to 2.84\% in the middle and 6.55\% in the final third. Despite fewer detected object mentions (1,734 to 1,150), hallucinated mentions increase from 26 to 73. The late-to-early difference reaches 5.12 percentage points (95\% CI: [2.88, 7.56]). All ten languages exhibit higher hallucination rates in the final third, although intermediate positions are not uniformly monotonic. Excluding 89 truncated responses preserves this trend (1.72\% $\rightarrow$ 2.79\% $\rightarrow$ 6.34\%). These findings provide direct output-level evidence of late-stage hallucination accumulation, consistent with autoregressive amplification.

\begin{table}[t]
\centering
\scriptsize
\setlength{\tabcolsep}{3pt}
\renewcommand{\arraystretch}{1.05}
\resizebox{\columnwidth}{!}{%
\begin{tabular}{lrrrrr}
\toprule
\textbf{Position}
& \textbf{Objects}
& \textbf{Hall.}
& \textbf{Rate (\%)}
& \textbf{Hall./100 tok.}
& \textbf{Resp. (\%)} \\
\midrule
Early  & 1,734 & 26 & 1.43 & 0.065 & 2.40 \\
Middle & 1,255 & 36 & 2.84 & 0.087 & 3.30 \\
Late   & 1,150 & 73 & \textbf{6.55} & \textbf{0.192} & \textbf{6.50} \\
\bottomrule
\end{tabular}%
}
\caption{Position-wise object hallucination in free-running AMBER descriptions. Objects and hallucinated mentions are pooled counts, whereas rates and token-normalized densities are macro-averaged across ten languages. Resp. denotes the percentage of responses containing at least one hallucinated object in the corresponding segment.}
\label{tab:position_hallucination}
\end{table}

\paragraph{Free-running versus teacher forcing: setup.}
Using the same 100-image set, we compare free-running and teacher-forced representation trajectories on InternVL3.5-14B-IT across English and the ten target languages. For teacher forcing, we reuse the canonical English descriptions and their ten semantically equivalent translations from Observation~3.
We analyze InternVL3.5-14B-IT over its primary layer range L14--27.
For both decoding conditions, hidden states are centered using the frozen
language-specific decoding centroid estimated during fitting.

In the free-running condition, each language generates greedily and
conditions on its own previously generated tokens.
In the teacher-forcing condition, the corresponding gold translated prefix
is supplied at every step, preventing previous model predictions from
entering subsequent decoding states.
Because sequence lengths and tokenizations differ across languages, we
compare trajectories at matched normalized generation progress
$q\in\{0,0.05,\ldots,1\}$ rather than at identical token indices.

For condition $c\in\{\mathrm{FR},\mathrm{TF}\}$, let
$t_{i,\ell,c}(q)$ denote the token position corresponding to progress $q$
for image $i$ and language $\ell$.
We center each representation as
\begin{equation}
\widetilde{\mathbf h}_{i,\ell,t}^{(k)}
=
\mathbf h_{i,\ell,t}^{(k)}
-
\boldsymbol{\mu}_{\ell,\mathrm{dec}}^{(k)},
\end{equation}
and define target-to-English trajectory divergence as
\begin{equation}
\begin{aligned}
D_{i,\ell,c}(q)
&=
\frac{1}{|\mathcal{A}|}
\sum_{k\in\mathcal{A}}
\Big[
1-
\cos\big(
\\[-2pt]
&\qquad
\widetilde{\mathbf h}_{i,\ell,t_{i,\ell,c}(q)}^{(k)},
\widetilde{\mathbf h}_{i,\mathrm{en},t_{i,\mathrm{en},c}(q)}^{(k)}
\big)
\Big],
\end{aligned}
\end{equation}
where $\mathcal{A}=\{14,\ldots,27\}$ is the set of analyzed layers.

To summarize trajectory evolution, we define
\begin{equation}
G_{i,\ell,c}
=
\overline{D}_{i,\ell,c}^{\,\mathrm{late}}
-
\overline{D}_{i,\ell,c}^{\,\mathrm{early}},
\end{equation}
where the early and late regions correspond to
$q\in[0,0.2]$ and $q\in[0.8,1.0]$, respectively.
The reported confidence intervals are obtained by paired bootstrap over
the free-running--teacher-forcing difference
\[
\Delta G_{i,\ell}
=
G_{i,\ell,\mathrm{FR}}
-
G_{i,\ell,\mathrm{TF}},
\]
using 10,000 bootstrap resamples.
The trajectory-wise confidence bands are computed analogously at each
progress point from
\[
\Delta D_{i,\ell,c}(q)
=
D_{i,\ell,c}(q)-D_{i,\ell,c}(0).
\]

\paragraph{Free-running versus teacher forcing: results.}
Figure~\ref{fig:trajectory_divergence} shows qualitatively different
trajectory dynamics under the two decoding protocols.
Free-running representations separate sharply once generation begins and
remain substantially more divergent, whereas teacher-forced trajectories
progressively reconverge.
The early-to-late divergence change is positive under free running and
negative under teacher forcing, with a paired difference of
$0.0831$ (95\% CI $[0.0745,0.0915]$).
The effect is positive for all ten target languages, and $71.1\%$ of
individual image--language pairs satisfy
$G_{\mathrm{FR}}>G_{\mathrm{TF}}$.

Free-running divergence does not increase monotonically:
self-generated prefixes induce a rapid early separation that is
subsequently sustained, while gold-prefix teacher forcing allows the
cross-lingual trajectories to reconverge.
The contrast remains significant after excluding both trajectory endpoints,
indicating that it is not driven solely by the initial state or
end-of-sequence behavior.
These results support the view that autoregressive feedback sustains and
amplifies otherwise recoverable cross-lingual deviations.

\paragraph{Budget-matched temporal intervention: setup.}
We next test whether the temporal placement of the correction matters
beyond its total perturbation magnitude.
AMBER contains 1,004 evaluation images; we reserve 200 images exclusively
for calibrating the energy-matching multipliers and evaluate the temporal
controls on the remaining 804 images.
Using the accumulated perturbation energy
\begin{equation}
E_{\Delta}
=
\sum_t
\sum_{k\in\mathcal{K}}
\left\|
\Delta\mathbf h_t^{(k)}
\right\|_2^2,
\end{equation}
we construct two temporally sparse controls whose mean perturbation energy
is matched to standard token-wise MVP-Steer.

For a temporal schedule $\mathcal T_c$,
\begin{equation}
\Delta\mathbf h_t^{(k)}
=
\begin{cases}
-c\,\rho_{\ell}w_k
\|\mathbf h_t^{(k)}\|_2
\mathbf u_t^{(k)},
& t\in\mathcal T_c,\\[2mm]
0,
& t\notin\mathcal T_c,
\end{cases}
\end{equation}
where $c$ is a schedule-specific global multiplier.
We compare standard token-wise intervention,
an initial-only intervention applied at $t=0$, and a periodic intervention
applied every four decoding tokens.
The sparse-schedule multipliers are selected only on the 200-image
calibration split so that their mean perturbation energy differs from the
repeated reference by at most $5\%$, and are then frozen before evaluation
on the disjoint 804-image test split.

\paragraph{Budget-matched temporal intervention: results.}
Table~\ref{tab:budget_matched_temporal} shows that total perturbation energy
alone does not explain the behavior of MVP-Steer.
The every-four-token schedule has essentially the same total energy and
incurs no clipping, yet provides substantially weaker hallucination
suppression than token-wise intervention.
Conversely, concentrating the same budget into the initial state recovers
comparable hallucination reduction only by entering a much larger
instantaneous perturbation regime, causing extensive clipping and a
significant loss of descriptive coverage.
Thus, repeated MVP-Steer realizes the same overall intervention budget as
frequent, conservative corrections rather than as sparse or concentrated
updates.

These analyses provide complementary evidence for autoregressive amplification. Object hallucinations concentrate in later generation positions, self-generated prefixes sustain excess cross-lingual divergence, and repeated conservative corrections outperform sparse budget-matched interventions without sacrificing descriptive coverage. These findings support the proposed butterfly-effect mechanism and motivate token-level correction throughout generation.

\section{Confidence Intervals}
\label{app:ci}

We report 95\% confidence intervals for the relative improvements in Table~\ref{tab:main_results} using a paired bootstrap over evaluation languages. Let $\bar{b}$ and $\bar{m}$ denote the macro-averaged scores of the baseline and the method over the evaluated languages. The relative improvement is
\begin{equation}
\Delta = s \cdot \frac{\bar{m} - \bar{b}}{\bar{b}} \times 100,
\label{eq:rel_improvement}
\end{equation}
where $s = +1$ for higher-is-better metrics and $s = -1$ for lower-is-better metrics, so that positive values always indicate improvement. In each bootstrap replicate, we sample ten languages with replacement from the ten evaluated languages and apply the same sampled indices to both the baseline and the method, preserving their pairing. We then recompute both macro averages and $\Delta$. We draw 200{,}000 replicates and report the 2.5th and 97.5th percentiles of the resulting distribution. Each method is paired with the baseline run that shares its decoding configuration: MVP-Steer, AdaIAT, and CLAIM with the default run of each backbone, OPERA with the InternVL3.5-14B-IT beam-search baseline, and VCD with its paired InternVL3.5-14B-IT rerun.

\section{Per-Language Results}
\label{app:Per-Language Results of MVP-Steer}

Tables~\ref{tab:amber_gemma}--\ref{tab:obj_qwen} provide the detailed per-language results of MVP-Steer reported in Table~\ref{tab:main_results}, covering all evaluated configurations for each backbone model. Table~\ref{tab:target_rho} summarizes the intervention strength $\rho$ applied to each language for each backbone model.

\begin{figure*}[htbp]
    \centering
    \includegraphics[width=1\linewidth]{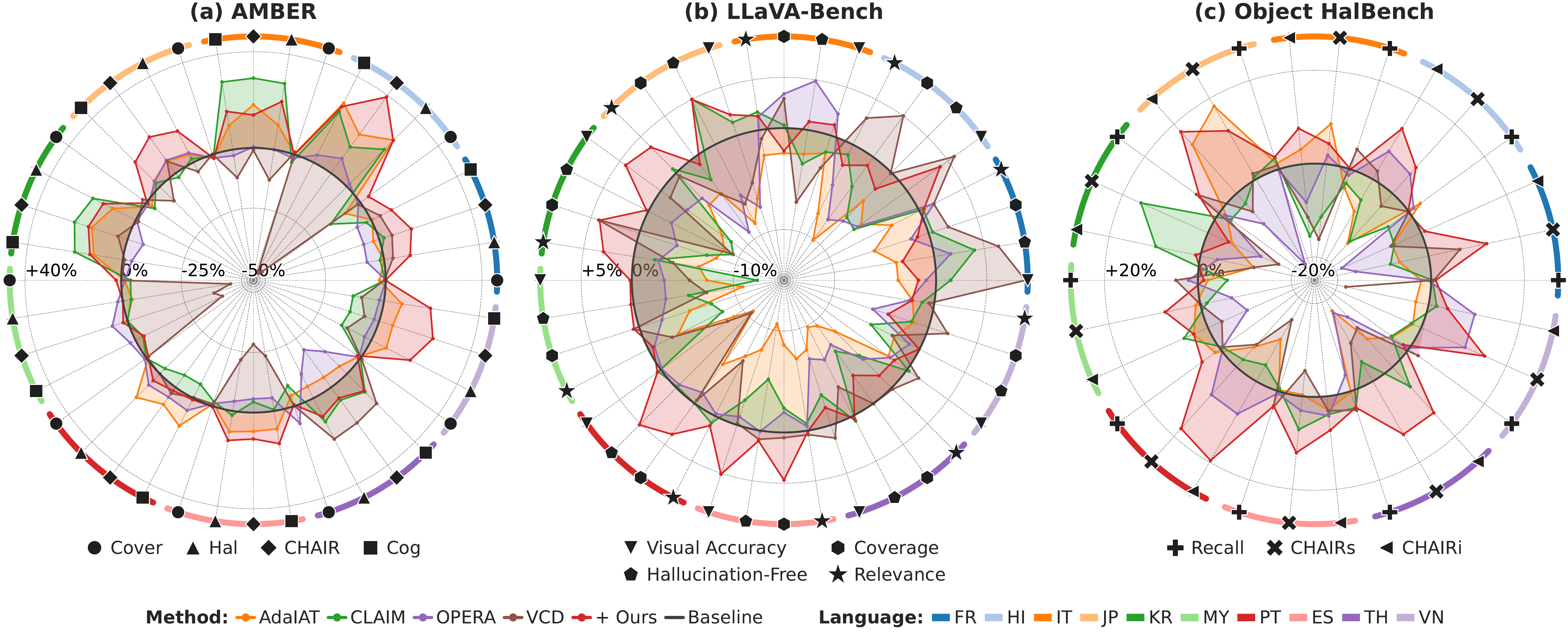}
    \caption{
    Per-language comparison of hallucination mitigation methods on
(a) AMBER, (b) LLaVA-Bench, and (c) Object HalBench.
Each sector corresponds to a target language, and radial axes denote the
evaluation metrics of each benchmark. Values show relative changes over the
corresponding baseline, with metric directions normalized such that outward
movement indicates improvement.
    }
    \label{fig:three_figures}
\end{figure*}
Figure~\ref{fig:three_figures} shows that MVP-Steer achieves broadly consistent hallucination reductions across languages while largely preserving generation quality and visual coverage. On InternVL3.5-14B-IT, AMBER CHAIR decreases in all ten languages, with reductions reaching 38.95\% in Hindi and 23.46\% in Vietnamese. On Object HalBench, CHAIR$_i$ also improves across all ten languages, reaching a 19.58\% reduction in Portuguese. In comparison, AdaIAT and CLAIM achieve some hallucination reductions but frequently compromise generation quality and object recall. VCD and OPERA exhibit less consistent improvements, with performance degradation across multiple languages and metrics. These results highlight MVP-Steer's more balanced performance across languages and benchmarks.

\begin{table*}[t]
\centering
\footnotesize
\setlength{\tabcolsep}{3pt}
\renewcommand{\arraystretch}{1.05}
\begin{tabular}{l|cccc}
\toprule
Language & Cover $\uparrow$ & CHAIR $\downarrow$ & Hal $\downarrow$ & Cog $\downarrow$ \\
\midrule
French & 60.51$\rightarrow$60.25 (-0.44\%) & 7.10$\rightarrow$6.67 (+6.00\%) & 22.91$\rightarrow$22.21 (+3.04\%) & 5.19$\rightarrow$4.96 (+4.41\%) \\
Hindi & 49.26$\rightarrow$49.31 (+0.11\%) & 4.23$\rightarrow$3.19 (+24.58\%) & 11.16$\rightarrow$9.06 (+18.75\%) & 2.41$\rightarrow$1.98 (+17.81\%) \\
Italian & 47.43$\rightarrow$47.56 (+0.29\%) & 7.84$\rightarrow$6.66 (+15.04\%) & 22.01$\rightarrow$19.62 (+10.86\%) & 4.89$\rightarrow$4.34 (+11.17\%) \\
Japanese & 53.15$\rightarrow$53.34 (+0.36\%) & 4.43$\rightarrow$4.16 (+6.18\%) & 12.55$\rightarrow$11.75 (+6.35\%) & 2.72$\rightarrow$2.50 (+8.17\%) \\
Korean & 53.72$\rightarrow$53.76 (+0.07\%) & 4.76$\rightarrow$4.36 (+8.40\%) & 15.74$\rightarrow$14.24 (+9.49\%) & 3.47$\rightarrow$3.12 (+10.10\%) \\
Malay & 63.27$\rightarrow$63.28 (+0.03\%) & 4.56$\rightarrow$4.45 (+2.40\%) & 15.64$\rightarrow$15.34 (+1.91\%) & 3.33$\rightarrow$3.29 (+1.05\%) \\
Portuguese & 55.33$\rightarrow$54.69 (-1.15\%) & 8.21$\rightarrow$7.06 (+13.98\%) & 26.00$\rightarrow$21.71 (+16.48\%) & 5.76$\rightarrow$4.78 (+16.93\%) \\
Spanish & 57.85$\rightarrow$57.39 (-0.79\%) & 8.58$\rightarrow$8.33 (+2.91\%) & 26.99$\rightarrow$26.89 (+0.37\%) & 6.22$\rightarrow$6.03 (+3.09\%) \\
Thai & 60.89$\rightarrow$60.78 (-0.18\%) & 4.76$\rightarrow$4.31 (+9.50\%) & 17.43$\rightarrow$15.04 (+13.71\%) & 3.98$\rightarrow$3.35 (+15.78\%) \\
Vietnamese & 63.90$\rightarrow$64.01 (+0.18\%) & 5.40$\rightarrow$5.39 (+0.25\%) & 20.22$\rightarrow$19.72 (+2.46\%) & 4.44$\rightarrow$4.39 (+1.20\%) \\
\bottomrule
\end{tabular}
\caption{AMBER results of Gemma-3-12B-IT across 10 languages on a 0--100 scale. Each entry reports Baseline $\rightarrow$ Ours (relative change).}
\label{tab:amber_gemma}
\end{table*}

\begin{table*}[t]
\centering
\footnotesize
\setlength{\tabcolsep}{3pt}
\renewcommand{\arraystretch}{1.05}

\sbox{\myfirsttablebox}{%
\begin{tabular}{l|cccc}
\hline
\textbf{Language} & \textbf{Visual Acc. $\uparrow$} & \textbf{Hall.-Free $\uparrow$} & \textbf{Coverage $\uparrow$} & \textbf{W/T/L} \\
\hline
French & 71.51$\rightarrow$74.87 (+4.71\%) & 66.09$\rightarrow$68.55 (+3.73\%) & 71.64$\rightarrow$75.01 (+4.70\%) & 18/34/6 \\
Hindi & 72.86$\rightarrow$73.74 (+1.22\%) & 64.13$\rightarrow$64.57 (+0.69\%) & 73.10$\rightarrow$73.47 (+0.51\%) & 7/42/9 \\
Italian & 73.52$\rightarrow$74.60 (+1.47\%) & 63.89$\rightarrow$64.98 (+1.71\%) & 72.54$\rightarrow$73.95 (+1.93\%) & 20/26/12 \\
Japanese & 68.21$\rightarrow$68.23 (+0.02\%) & 59.50$\rightarrow$60.78 (+2.15\%) & 70.65$\rightarrow$68.62 (-2.88\%) & 14/30/14 \\
Korean & 71.48$\rightarrow$72.35 (+1.22\%) & 66.82$\rightarrow$65.27 (-2.32\%) & 73.28$\rightarrow$73.90 (+0.84\%) & 16/21/21 \\
Malay & 73.57$\rightarrow$74.20 (+0.86\%) & 64.41$\rightarrow$66.90 (+3.87\%) & 72.25$\rightarrow$73.09 (+1.16\%) & 18/29/11 \\
Portuguese & 74.85$\rightarrow$73.98 (-1.17\%) & 66.25$\rightarrow$67.13 (+1.33\%) & 74.22$\rightarrow$74.03 (-0.25\%) & 15/29/14 \\
Spanish & 73.11$\rightarrow$72.70 (-0.56\%) & 65.64$\rightarrow$63.65 (-3.04\%) & 74.73$\rightarrow$74.84 (+0.14\%) & 8/35/15 \\
Thai & 68.96$\rightarrow$67.05 (-2.77\%) & 64.04$\rightarrow$64.43 (+0.61\%) & 70.24$\rightarrow$67.61 (-3.74\%) & 21/18/19 \\
Vietnamese & 76.38$\rightarrow$77.28 (+1.18\%) & 74.21$\rightarrow$74.53 (+0.44\%) & 74.82$\rightarrow$75.14 (+0.43\%) & 10/42/6 \\
\hline
\end{tabular}%
}
\usebox{\myfirsttablebox}

\smallskip

\begin{tabular*}{\wd\myfirsttablebox}{l|@{\extracolsep{\fill}}ccc}
\hline
\textbf{Language} & \textbf{Relevance $\uparrow$} & \textbf{Lang. Adh. $\uparrow$} & \textbf{Overall $\uparrow$} \\
\hline
French & 90.59$\rightarrow$91.15 (+0.62\%) & 99.49$\rightarrow$100.00 (+0.52\%) & 67.14$\rightarrow$69.84 (+4.01\%) \\
Hindi & 91.20$\rightarrow$91.39 (+0.21\%) & 98.57$\rightarrow$97.93 (-0.64\%) & 68.74$\rightarrow$68.83 (+0.14\%) \\
Italian & 89.78$\rightarrow$92.04 (+2.51\%) & 98.72$\rightarrow$100.00 (+1.30\%) & 66.83$\rightarrow$69.04 (+3.30\%) \\
Japanese & 90.30$\rightarrow$90.95 (+0.72\%) & 99.61$\rightarrow$100.00 (+0.39\%) & 63.26$\rightarrow$63.85 (+0.94\%) \\
Korean & 90.42$\rightarrow$92.84 (+2.68\%) & 99.74$\rightarrow$100.00 (+0.26\%) & 67.68$\rightarrow$67.53 (-0.22\%) \\
Malay & 89.75$\rightarrow$90.84 (+1.22\%) & 96.28$\rightarrow$96.52 (+0.25\%) & 67.48$\rightarrow$68.44 (+1.42\%) \\
Portuguese & 91.63$\rightarrow$92.21 (+0.63\%) & 99.56$\rightarrow$100.00 (+0.45\%) & 69.38$\rightarrow$68.73 (-0.94\%) \\
Spanish & 90.73$\rightarrow$89.94 (-0.86\%) & 99.49$\rightarrow$100.00 (+0.52\%) & 69.26$\rightarrow$68.30 (-1.38\%) \\
Thai & 91.82$\rightarrow$91.97 (+0.17\%) & 99.49$\rightarrow$100.00 (+0.52\%) & 65.26$\rightarrow$63.82 (-2.20\%) \\
Vietnamese & 93.71$\rightarrow$93.97 (+0.27\%) & 100.00$\rightarrow$100.00 (0.00\%) & 73.09$\rightarrow$73.70 (+0.84\%) \\
\hline
\end{tabular*}

\caption{LLaVA-Bench results of Gemma-3-12B-IT across 10 languages on a 0--100 scale. Each entry reports Baseline $\rightarrow$ Ours (relative change).}
\label{tab:llava_gemma}
\end{table*}

\begin{table*}[t]
\centering
\footnotesize
\setlength{\tabcolsep}{3pt}
\renewcommand{\arraystretch}{1.05}
\begin{tabular}{l|ccc}
\toprule
Language & CHAIR$_s$ $\downarrow$ & CHAIR$_i$ $\downarrow$ & Object Recall $\uparrow$ \\
\midrule
French & 14.97$\rightarrow$14.63 (+2.22\%) & 7.96$\rightarrow$7.82 (+1.74\%) & 61.33$\rightarrow$58.84 (-4.07\%) \\
Hindi & 14.38$\rightarrow$14.38 (0.00\%) & 7.81$\rightarrow$7.40 (+5.33\%) & 61.33$\rightarrow$62.47 (+1.86\%) \\
Italian & 12.80$\rightarrow$13.06 (-2.00\%) & 7.00$\rightarrow$7.19 (-2.66\%) & 59.36$\rightarrow$59.04 (-0.53\%) \\
Japanese & 17.01$\rightarrow$14.48 (+14.88\%) & 8.92$\rightarrow$7.80 (+12.50\%) & 59.46$\rightarrow$60.19 (+1.22\%) \\
Korean & 12.94$\rightarrow$10.65 (+17.66\%) & 6.51$\rightarrow$5.55 (+14.81\%) & 58.21$\rightarrow$60.19 (+3.39\%) \\
Malay & 12.03$\rightarrow$10.07 (+16.28\%) & 6.23$\rightarrow$5.74 (+7.80\%) & 57.90$\rightarrow$58.00 (+0.18\%) \\
Portuguese & 15.86$\rightarrow$14.83 (+6.52\%) & 8.45$\rightarrow$7.72 (+8.71\%) & 59.67$\rightarrow$60.91 (+2.09\%) \\
Spanish & 15.81$\rightarrow$15.36 (+2.84\%) & 8.07$\rightarrow$8.22 (-1.89\%) & 62.79$\rightarrow$62.68 (-0.17\%) \\
Thai & 13.54$\rightarrow$10.38 (+23.34\%) & 7.66$\rightarrow$5.38 (+29.73\%) & 60.19$\rightarrow$62.16 (+3.28\%) \\
Vietnamese & 13.94$\rightarrow$13.89 (+0.35\%) & 7.76$\rightarrow$7.02 (+9.46\%) & 56.86$\rightarrow$57.80 (+1.65\%) \\
\bottomrule
\end{tabular}
\caption{Object HalBench results of Gemma-3-12B-IT across 10 languages on a 0--100 scale. Each entry reports Baseline $\rightarrow$ Ours (relative change).}
\label{tab:obj_gemma}
\end{table*}

\begin{table*}[t]
\centering
\footnotesize
\setlength{\tabcolsep}{3pt}
\renewcommand{\arraystretch}{1.05}
\begin{tabular}{l|cccc}
\toprule
Language & Cover $\uparrow$ & CHAIR $\downarrow$ & Hal $\downarrow$ & Cog $\downarrow$ \\
\midrule
French & 55.81$\rightarrow$55.72 (-0.16\%) & 4.27$\rightarrow$3.67 (+14.00\%) & 12.75$\rightarrow$11.35 (+10.94\%) & 2.77$\rightarrow$2.51 (+9.17\%) \\
Hindi & 33.45$\rightarrow$34.85 (+4.19\%) & 2.76$\rightarrow$1.68 (+38.95\%) & 5.88$\rightarrow$4.28 (+27.12\%) & 1.23$\rightarrow$0.91 (+25.95\%) \\
Italian & 42.83$\rightarrow$42.77 (-0.13\%) & 3.79$\rightarrow$3.27 (+13.79\%) & 10.36$\rightarrow$8.27 (+20.19\%) & 2.18$\rightarrow$1.83 (+15.99\%) \\
Japanese & 51.05$\rightarrow$50.18 (-1.72\%) & 4.30$\rightarrow$3.50 (+18.46\%) & 12.25$\rightarrow$10.46 (+14.63\%) & 2.62$\rightarrow$2.24 (+14.38\%) \\
Korean & 46.79$\rightarrow$46.61 (-0.40\%) & 5.60$\rightarrow$4.64 (+17.15\%) & 15.84$\rightarrow$13.45 (+15.09\%) & 3.61$\rightarrow$3.11 (+13.81\%) \\
Malay & 49.03$\rightarrow$50.07 (+2.13\%) & 4.12$\rightarrow$4.04 (+2.05\%) & 10.66$\rightarrow$10.66 (0.00\%) & 2.28$\rightarrow$2.37 (-3.92\%) \\
Portuguese & 51.25$\rightarrow$50.79 (-0.89\%) & 4.47$\rightarrow$4.39 (+1.93\%) & 12.65$\rightarrow$12.15 (+3.94\%) & 2.78$\rightarrow$2.76 (+0.54\%) \\
Spanish & 56.49$\rightarrow$56.63 (+0.25\%) & 4.74$\rightarrow$4.22 (+11.00\%) & 16.14$\rightarrow$14.14 (+12.35\%) & 3.57$\rightarrow$3.08 (+13.58\%) \\
Thai & 37.10$\rightarrow$36.86 (-0.63\%) & 5.78$\rightarrow$5.46 (+5.50\%) & 15.14$\rightarrow$13.84 (+8.55\%) & 3.39$\rightarrow$3.05 (+10.13\%) \\
Vietnamese & 60.87$\rightarrow$59.88 (-1.63\%) & 5.41$\rightarrow$4.14 (+23.46\%) & 18.13$\rightarrow$14.84 (+18.13\%) & 3.99$\rightarrow$3.21 (+19.63\%) \\
\bottomrule
\end{tabular}
\caption{AMBER results of InternVL3.5-14B-IT across 10 languages on a 0--100 scale. Each entry reports Baseline $\rightarrow$ Ours (relative change).}
\label{tab:amber_internvl}
\end{table*}

\begin{table*}[t]
\centering
\footnotesize
\setlength{\tabcolsep}{3pt}
\renewcommand{\arraystretch}{1.05}

\sbox{\myfirsttablebox}{%
\begin{tabular}{l|cccc}
\hline
\textbf{Language} & \textbf{Visual Acc. $\uparrow$} & \textbf{Hall.-Free $\uparrow$} & \textbf{Coverage $\uparrow$} & \textbf{W/T/L} \\
\hline
French & 64.99$\rightarrow$63.86 (-1.74\%) & 66.54$\rightarrow$64.43 (-3.16\%) & 65.67$\rightarrow$64.92 (-1.15\%) & 21/20/17 \\
Hindi & 46.97$\rightarrow$48.87 (+4.05\%) & 55.15$\rightarrow$53.89 (-2.28\%) & 43.80$\rightarrow$43.36 (-1.01\%) & 16/33/9 \\
Italian & 60.55$\rightarrow$61.21 (+1.10\%) & 63.63$\rightarrow$64.16 (+0.83\%) & 65.30$\rightarrow$63.83 (-2.25\%) & 19/20/19 \\
Japanese & 57.84$\rightarrow$59.09 (+2.16\%) & 55.29$\rightarrow$58.06 (+5.01\%) & 60.50$\rightarrow$59.95 (-0.90\%) & 21/25/12 \\
Korean & 54.52$\rightarrow$56.86 (+4.30\%) & 56.68$\rightarrow$56.81 (+0.22\%) & 55.68$\rightarrow$57.97 (+4.12\%) & 24/16/18 \\
Malay & 59.89$\rightarrow$59.98 (+0.17\%) & 59.80$\rightarrow$59.96 (+0.28\%) & 59.52$\rightarrow$59.85 (+0.54\%) & 22/19/17 \\
Portuguese & 61.73$\rightarrow$61.98 (+0.42\%) & 59.64$\rightarrow$62.71 (+5.15\%) & 62.09$\rightarrow$64.42 (+3.77\%) & 18/22/18 \\
Spanish & 65.21$\rightarrow$68.54 (+5.10\%) & 64.83$\rightarrow$65.50 (+1.03\%) & 64.77$\rightarrow$67.81 (+4.69\%) & 20/23/15 \\
Thai & 54.04$\rightarrow$53.07 (-1.81\%) & 53.43$\rightarrow$53.61 (+0.34\%) & 53.75$\rightarrow$51.92 (-3.41\%) & 21/21/16 \\
Vietnamese & 68.17$\rightarrow$67.09 (-1.58\%) & 64.99$\rightarrow$64.96 (-0.05\%) & 69.76$\rightarrow$66.84 (-4.19\%) & 11/30/17 \\
\hline
\end{tabular}%
}
\usebox{\myfirsttablebox}

\smallskip

\begin{tabular*}{\wd\myfirsttablebox}{l|@{\extracolsep{\fill}}ccc}
\hline
\textbf{Language} & \textbf{Relevance $\uparrow$} & \textbf{Lang. Adh. $\uparrow$} & \textbf{Overall $\uparrow$} \\
\hline
French & 82.24$\rightarrow$82.65 (+0.50\%) & 88.81$\rightarrow$92.50 (+4.16\%) & 61.27$\rightarrow$61.83 (+0.92\%) \\
Hindi & 62.80$\rightarrow$61.38 (-2.26\%) & 88.36$\rightarrow$87.52 (-0.95\%) & 42.43$\rightarrow$42.09 (-0.80\%) \\
Italian & 79.56$\rightarrow$80.65 (+1.38\%) & 91.83$\rightarrow$92.70 (+0.95\%) & 60.20$\rightarrow$59.63 (-0.94\%) \\
Japanese & 74.06$\rightarrow$76.70 (+3.55\%) & 71.98$\rightarrow$73.03 (+1.46\%) & 50.86$\rightarrow$52.41 (+3.05\%) \\
Korean & 74.62$\rightarrow$76.88 (+3.02\%) & 81.58$\rightarrow$82.83 (+1.53\%) & 50.87$\rightarrow$52.84 (+3.87\%) \\
Malay & 76.98$\rightarrow$76.59 (-0.51\%) & 73.16$\rightarrow$75.95 (+3.80\%) & 53.78$\rightarrow$54.79 (+1.89\%) \\
Portuguese & 77.99$\rightarrow$78.84 (+1.09\%) & 90.24$\rightarrow$87.13 (-3.45\%) & 57.43$\rightarrow$58.51 (+1.89\%) \\
Spanish & 83.95$\rightarrow$84.02 (+0.08\%) & 92.96$\rightarrow$92.05 (-0.97\%) & 61.95$\rightarrow$64.53 (+4.16\%) \\
Thai & 68.70$\rightarrow$67.52 (-1.72\%) & 41.85$\rightarrow$47.51 (+13.53\%) & 42.95$\rightarrow$43.28 (+0.77\%) \\
Vietnamese & 86.09$\rightarrow$84.16 (-2.24\%) & 94.41$\rightarrow$93.92 (-0.53\%) & 66.08$\rightarrow$64.16 (-2.91\%) \\
\hline
\end{tabular*}

\caption{LLaVA-Bench results of InternVL3.5-14B-IT across 10 languages on a 0--100 scale. Each entry reports Baseline $\rightarrow$ Ours (relative change).}
\label{tab:llava_internvl}
\end{table*}

\begin{table*}[t]
\centering
\footnotesize
\setlength{\tabcolsep}{3pt}
\renewcommand{\arraystretch}{1.05}
\begin{tabular}{l|ccc}
\toprule
Language & CHAIR$_s$ $\downarrow$ & CHAIR$_i$ $\downarrow$ & Object Recall $\uparrow$ \\
\midrule
French & 22.68$\rightarrow$19.80 (+12.72\%) & 10.80$\rightarrow$10.71 (+0.87\%) & 63.51$\rightarrow$64.14 (+0.98\%) \\
Hindi & 17.44$\rightarrow$16.13 (+7.50\%) & 11.00$\rightarrow$9.63 (+12.51\%) & 47.92$\rightarrow$47.82 (-0.22\%) \\
Italian & 22.15$\rightarrow$21.16 (+4.45\%) & 11.76$\rightarrow$10.85 (+7.70\%) & 63.20$\rightarrow$63.20 (0.00\%) \\
Japanese & 15.28$\rightarrow$13.45 (+11.98\%) & 8.80$\rightarrow$7.24 (+17.80\%) & 57.07$\rightarrow$58.63 (+2.73\%) \\
Korean & 22.11$\rightarrow$23.18 (-4.86\%) & 11.82$\rightarrow$11.70 (+1.03\%) & 55.82$\rightarrow$58.84 (+5.40\%) \\
Malay & 17.87$\rightarrow$16.49 (+7.71\%) & 9.93$\rightarrow$9.60 (+3.34\%) & 57.48$\rightarrow$56.76 (-1.27\%) \\
Portuguese & 26.46$\rightarrow$21.77 (+17.73\%) & 13.70$\rightarrow$11.01 (+19.58\%) & 60.91$\rightarrow$63.83 (+4.78\%) \\
Spanish & 22.07$\rightarrow$19.39 (+12.15\%) & 11.47$\rightarrow$10.63 (+7.31\%) & 62.58$\rightarrow$64.66 (+3.32\%) \\
Thai & 14.29$\rightarrow$12.41 (+13.12\%) & 8.36$\rightarrow$7.26 (+13.21\%) & 51.25$\rightarrow$53.12 (+3.65\%) \\
Vietnamese & 24.32$\rightarrow$20.69 (+14.91\%) & 11.87$\rightarrow$11.37 (+4.16\%) & 63.31$\rightarrow$62.37 (-1.48\%) \\
\bottomrule
\end{tabular}
\caption{Object HalBench results of InternVL3.5-14B-IT across 10 languages on a 0--100 scale. Each entry reports Baseline $\rightarrow$ Ours (relative change).}
\label{tab:obj_internvl}
\end{table*}

\begin{table*}[t]
\centering
\footnotesize
\setlength{\tabcolsep}{3pt}
\renewcommand{\arraystretch}{1.05}
\begin{tabular}{l|cccc}
\toprule
Language & Cover $\uparrow$ & CHAIR $\downarrow$ & Hal $\downarrow$ & Cog $\downarrow$ \\
\midrule
French & 65.12$\rightarrow$65.23 (+0.17\%) & 4.84$\rightarrow$4.54 (+6.23\%) & 17.53$\rightarrow$17.23 (+1.70\%) & 4.07$\rightarrow$3.94 (+3.22\%) \\
Hindi & 40.44$\rightarrow$40.55 (+0.26\%) & 2.69$\rightarrow$2.39 (+11.35\%) & 6.57$\rightarrow$5.98 (+9.09\%) & 1.42$\rightarrow$1.28 (+9.82\%) \\
Italian & 52.21$\rightarrow$51.72 (-0.95\%) & 6.39$\rightarrow$6.04 (+5.46\%) & 19.22$\rightarrow$18.43 (+4.15\%) & 4.32$\rightarrow$4.04 (+6.46\%) \\
Japanese & 59.75$\rightarrow$59.56 (-0.31\%) & 6.70$\rightarrow$6.28 (+6.17\%) & 21.12$\rightarrow$20.92 (+0.94\%) & 4.76$\rightarrow$4.74 (+0.45\%) \\
Korean & 58.50$\rightarrow$57.87 (-1.08\%) & 5.95$\rightarrow$5.75 (+3.31\%) & 21.12$\rightarrow$19.72 (+6.60\%) & 4.77$\rightarrow$4.50 (+5.54\%) \\
Malay & 62.38$\rightarrow$62.30 (-0.14\%) & 6.14$\rightarrow$5.79 (+5.72\%) & 21.12$\rightarrow$20.32 (+3.77\%) & 4.78$\rightarrow$4.48 (+6.25\%) \\
Portuguese & 58.82$\rightarrow$59.28 (+0.78\%) & 5.84$\rightarrow$5.47 (+6.39\%) & 19.62$\rightarrow$19.12 (+2.54\%) & 4.37$\rightarrow$4.17 (+4.78\%) \\
Spanish & 61.74$\rightarrow$61.44 (-0.49\%) & 6.60$\rightarrow$6.28 (+4.80\%) & 23.31$\rightarrow$22.01 (+5.56\%) & 5.12$\rightarrow$4.91 (+3.96\%) \\
Thai & 62.38$\rightarrow$62.74 (+0.58\%) & 6.92$\rightarrow$6.33 (+8.51\%) & 25.70$\rightarrow$23.31 (+9.30\%) & 5.82$\rightarrow$5.22 (+10.32\%) \\
Vietnamese & 66.09$\rightarrow$65.70 (-0.59\%) & 8.03$\rightarrow$7.02 (+12.54\%) & 28.88$\rightarrow$24.90 (+13.79\%) & 6.82$\rightarrow$5.76 (+15.56\%) \\
\bottomrule
\end{tabular}
\caption{AMBER results of Qwen3-VL-8B-IT across 10 languages on a 0--100 scale. Each entry reports Baseline $\rightarrow$ Ours (relative change).}
\label{tab:amber_qwen}
\end{table*}

\begin{table*}[t]
\centering
\footnotesize
\setlength{\tabcolsep}{3pt}
\renewcommand{\arraystretch}{1.05}

\sbox{\myfirsttablebox}{%
\begin{tabular}{l|cccc}
\hline
\textbf{Language} & \textbf{Visual Acc. $\uparrow$} & \textbf{Hall.-Free $\uparrow$} & \textbf{Coverage $\uparrow$} & \textbf{W/T/L} \\
\hline
French & 68.08$\rightarrow$69.36 (+1.88\%) & 58.49$\rightarrow$59.71 (+2.07\%) & 65.92$\rightarrow$68.23 (+3.50\%) & 18/27/13 \\
Hindi & 62.71$\rightarrow$60.83 (-3.00\%) & 57.78$\rightarrow$55.01 (-4.79\%) & 51.75$\rightarrow$50.56 (-2.30\%) & 14/24/20 \\
Italian & 67.86$\rightarrow$70.77 (+4.30\%) & 56.02$\rightarrow$57.79 (+3.16\%) & 66.54$\rightarrow$69.76 (+4.85\%) & 22/20/16 \\
Japanese & 66.62$\rightarrow$65.67 (-1.43\%) & 52.61$\rightarrow$54.53 (+3.64\%) & 63.76$\rightarrow$61.90 (-2.92\%) & 19/22/17 \\
Korean & 61.91$\rightarrow$66.30 (+7.10\%) & 48.90$\rightarrow$52.45 (+7.27\%) & 57.21$\rightarrow$61.73 (+7.92\%) & 24/16/18 \\
Malay & 68.74$\rightarrow$68.35 (-0.57\%) & 58.20$\rightarrow$57.91 (-0.49\%) & 66.89$\rightarrow$66.98 (+0.12\%) & 15/26/17 \\
Portuguese & 68.81$\rightarrow$68.81 (0.00\%) & 57.55$\rightarrow$59.21 (+2.88\%) & 66.15$\rightarrow$64.89 (-1.91\%) & 12/34/12 \\
Spanish & 67.36$\rightarrow$67.28 (-0.13\%) & 57.97$\rightarrow$57.61 (-0.62\%) & 66.43$\rightarrow$67.22 (+1.19\%) & 12/31/15 \\
Thai & 60.24$\rightarrow$59.16 (-1.79\%) & 53.12$\rightarrow$54.92 (+3.37\%) & 55.89$\rightarrow$54.66 (-2.20\%) & 18/17/23 \\
Vietnamese & 64.99$\rightarrow$67.55 (+3.93\%) & 54.29$\rightarrow$58.12 (+7.04\%) & 64.31$\rightarrow$64.38 (+0.10\%) & 22/17/19 \\
\hline
\end{tabular}%
}
\usebox{\myfirsttablebox}

\smallskip

\begin{tabular*}{\wd\myfirsttablebox}{l|@{\extracolsep{\fill}}ccc}
\hline
\textbf{Language} & \textbf{Relevance $\uparrow$} & \textbf{Lang. Adh. $\uparrow$} & \textbf{Overall $\uparrow$} \\
\hline
French & 88.34$\rightarrow$89.43 (+1.24\%) & 99.04$\rightarrow$99.23 (+0.19\%) & 62.23$\rightarrow$64.08 (+2.98\%) \\
Hindi & 77.33$\rightarrow$75.97 (-1.76\%) & 95.71$\rightarrow$95.83 (+0.13\%) & 53.02$\rightarrow$50.46 (-4.83\%) \\
Italian & 86.10$\rightarrow$87.22 (+1.31\%) & 98.45$\rightarrow$97.93 (-0.52\%) & 60.62$\rightarrow$62.97 (+3.87\%) \\
Japanese & 80.53$\rightarrow$81.94 (+1.76\%) & 99.10$\rightarrow$98.70 (-0.40\%) & 57.65$\rightarrow$56.69 (-1.67\%) \\
Korean & 79.74$\rightarrow$81.01 (+1.59\%) & 96.83$\rightarrow$97.41 (+0.60\%) & 52.14$\rightarrow$56.27 (+7.92\%) \\
Malay & 87.15$\rightarrow$87.46 (+0.35\%) & 85.17$\rightarrow$86.74 (+1.85\%) & 61.38$\rightarrow$61.30 (-0.14\%) \\
Portuguese & 86.05$\rightarrow$85.04 (-1.17\%) & 99.74$\rightarrow$99.04 (-0.70\%) & 62.80$\rightarrow$62.65 (-0.25\%) \\
Spanish & 86.92$\rightarrow$86.07 (-0.98\%) & 100.00$\rightarrow$99.56 (-0.44\%) & 61.09$\rightarrow$60.63 (-0.75\%) \\
Thai & 78.81$\rightarrow$77.92 (-1.13\%) & 98.70$\rightarrow$98.97 (+0.28\%) & 53.14$\rightarrow$52.84 (-0.55\%) \\
Vietnamese & 84.48$\rightarrow$84.75 (+0.32\%) & 99.22$\rightarrow$100.00 (+0.79\%) & 59.35$\rightarrow$61.42 (+3.49\%) \\
\hline
\end{tabular*}

\caption{LLaVA-Bench results of Qwen3-VL-8B-IT across 10 languages on a 0--100 scale. Each entry reports Baseline $\rightarrow$ Ours (relative change).}
\label{tab:llava_qwen}
\end{table*}

\begin{table*}[t]
\centering
\footnotesize
\setlength{\tabcolsep}{3pt}
\renewcommand{\arraystretch}{1.05}
\begin{tabular}{l|ccc}
\toprule
Language & CHAIR$_s$ $\downarrow$ & CHAIR$_i$ $\downarrow$ & Object Recall $\uparrow$ \\
\midrule
French & 10.99$\rightarrow$8.90 (+19.07\%) & 6.20$\rightarrow$5.27 (+14.99\%) & 51.87$\rightarrow$52.29 (+0.80\%) \\
Hindi & 5.22$\rightarrow$4.51 (+13.65\%) & 3.46$\rightarrow$2.99 (+13.55\%) & 37.73$\rightarrow$37.11 (-1.65\%) \\
Italian & 11.19$\rightarrow$9.57 (+14.43\%) & 6.13$\rightarrow$5.60 (+8.68\%) & 50.94$\rightarrow$50.83 (-0.20\%) \\
Japanese & 7.83$\rightarrow$8.45 (-7.94\%) & 4.48$\rightarrow$4.84 (-7.99\%) & 48.75$\rightarrow$49.06 (+0.64\%) \\
Korean & 11.23$\rightarrow$10.51 (+6.45\%) & 6.79$\rightarrow$6.86 (-0.98\%) & 45.63$\rightarrow$46.57 (+2.05\%) \\
Malay & 9.52$\rightarrow$9.45 (+0.73\%) & 5.79$\rightarrow$5.65 (+2.34\%) & 49.06$\rightarrow$50.31 (+2.54\%) \\
Portuguese & 12.14$\rightarrow$9.39 (+22.70\%) & 7.13$\rightarrow$5.64 (+20.86\%) & 51.46$\rightarrow$50.42 (-2.02\%) \\
Spanish & 9.93$\rightarrow$10.03 (-1.04\%) & 5.28$\rightarrow$5.75 (-8.85\%) & 55.93$\rightarrow$56.24 (+0.56\%) \\
Thai & 8.21$\rightarrow$7.46 (+9.09\%) & 5.65$\rightarrow$4.68 (+17.18\%) & 41.68$\rightarrow$44.49 (+6.73\%) \\
Vietnamese & 12.20$\rightarrow$12.63 (-3.58\%) & 7.49$\rightarrow$7.95 (-6.22\%) & 53.95$\rightarrow$56.55 (+4.82\%) \\
\bottomrule
\end{tabular}
\caption{Object HalBench results of Qwen3-VL-8B-IT across 10 languages on a 0--100 scale. Each entry reports Baseline $\rightarrow$ Ours (relative change).}
\label{tab:obj_qwen}
\end{table*}

\begin{table*}[htbp]
\centering
\small
\setlength{\tabcolsep}{6pt}
\renewcommand{\arraystretch}{1.05}
\begin{tabular}{l|ccc}
\toprule
Language
& Gemma-3-12B-IT
& Qwen3-VL-8B-IT
& InternVL3.5-14B-IT \\
\midrule
French     & 0.00025  & 0.001  & 0.010  \\
Hindi      & 0.00025  & 0.0025 & 0.010  \\
Italian    & 0.001    & 0.005  & 0.005  \\
Japanese   & 0.0005   & 0.0015 & 0.0005 \\
Korean     & 0.0015   & 0.005  & 0.010  \\
Malay      & 0.00025 & 0.004  & 0.0005 \\
Portuguese & 0.0015  & 0.0005 & 0.005  \\
Spanish    & 0.00025  & 0.0025 & 0.005  \\
Thai       & 0.0015   & 0.015  & 0.0015 \\
Vietnamese & 0.00005  & 0.010  & 0.010  \\
\bottomrule
\end{tabular}
\caption{Target intervention strength $\rho$ for each language across three backbone models.}
\label{tab:target_rho}
\end{table*}

\section{Permutation-Based Rank Selection}
\label{app:rank}

This section details the rank selection used in Section~\ref{sec:method_gcca}.
We determine $r$ without using downstream hallucination benchmarks. The aligned corpus is divided equally into fitting and validation partitions using a fixed split. GCCA is estimated on the fitting partition with a maximum candidate rank of $64$.

For validation sample representations, let
$\mathbf{g}_{\ell}=\mathbf{Z}_{\ell}\mathbf{W}_{\ell}$
denote the language-specific realization of the learned common latent representation. For each dimension $q$, we compute its weakest cross-language agreement:
\begin{equation}
    c_{q}^{\min}
    =
    \min_{\ell\neq m}
    \operatorname{corr}
    \left(
        g_{\ell,q},
        g_{m,q}
    \right).
\end{equation}

To determine whether this agreement exceeds chance alignment, we independently permute the sample correspondence within each language. For each of $32$ permutations, we record the largest absolute correlation observed across all candidate dimensions and language pairs. Let $\tau_{\mathrm{null}}$ be the $95$th percentile of these permutation maxima. We retain the longest leading sequence of GCCA dimensions satisfying
\begin{equation}
    c_{q}^{\min}
    >
    \tau_{\mathrm{null}}.
\end{equation}

After selecting the rank, we refit
$\boldsymbol{\mu}_{\ell,p}^{(k)}$
and
$\mathbf{B}_{\ell,p}^{(k)}$
using all $4{,}146$ aligned samples. The selected rank is shared across languages for a given phase and layer, whereas the hidden-space basis remains language-conditioned.

\section{Visual Rank Selection}
\label{app:visual_rank}
This section details the selection of the visual rank $m_{p,k}$ in Section~\ref{sec:method_visual}. In implementation, the visual basis $\mathbf{U}$ is obtained by uncentered randomized PCA
  over the pooled counterfactual differences. For rank selection, the 512
  examples of each language are split into 384 fitting and 128 held-out examples,
  and candidate ranks are $m\in\{4,8,16,32,64\}$. For a difference matrix
  $\mathbf D$ and the first $m$ principal directions $\mathbf U_m$, we measure
  the retained energy as
  \begin{equation}
  E_m(\mathbf D)
  =
  \frac{\|\mathbf D\mathbf U_m^\top\|_F^2}
       {\|\mathbf D\|_F^2}.
  \end{equation}
  After removing the isotropic random-subspace expectation $m/d$, retention is
  measured relative to the rank-64 basis:
  \begin{equation}
  R_m(\mathbf D)
  =
  \frac{\max(E_m(\mathbf D)-m/d,0)}
       {\max(E_{64}(\mathbf D)-64/d,\epsilon)}.
  \end{equation}
  We select the smallest rank whose retention is at least $0.90$ on the pooled
  held-out differences and on the median language, at least $0.75$ on the
  worst-case language, and at least $0.85$ under median
  leave-one-language-out evaluation. If no candidate satisfies these criteria,
  we use rank 64. Finally, we refit the uncentered PCA on all 512 examples from
  all languages and retain the selected number of principal directions.

\section{Why Direct Sequential Projection Is Insufficient}
\label{app:seq_proj}

Consider the centered state
\begin{equation}
    \boldsymbol{\delta}
    =
    \mathbf{h}
    -
    \boldsymbol{\mu}_{\ell}
\end{equation}
and its shared-space complement
\begin{equation}
    \mathbf{c}
    =
    (\mathbf{I}-\mathbf{P}_{s})
    \boldsymbol{\delta}.
\end{equation}
A straightforward strategy would protect the raw visual projection by forming
\begin{equation}
    \mathbf{r}_{\mathrm{seq}}
    =
    (\mathbf{I}-\gamma\mathbf{P}_{u})
    \mathbf{c},
    \qquad
    \mathbf{P}_{u}
    =
    \mathbf{U}^{\top}\mathbf{U}.
\end{equation}
However, $\mathbf{U}$ may overlap with the shared basis $\mathbf{B}$. Consequently, $\mathbf{P}_{u}$ and $\mathbf{P}_{s}$ need not commute, and the projected term
$\mathbf{P}_{u}\mathbf{c}$ can itself contain components in the shared subspace. Subtracting this term can therefore perturb information that the shared-space decomposition is intended to preserve. The orthogonalized visual basis in Section~\ref{sec:method_strict} avoids this, since it is orthogonal to the shared subspace by construction.

\section{Depth-Dependent Layer Weighting}
\label{app:layer_weight}

Rather than applying the same intervention strength to every selected decoder layer, we use a smooth depth-dependent weighting scheme that emphasizes a central intervention region while gradually reducing the strength near its boundaries.

Let $\mathcal{K}$ denote the set of intervened layers. We first define the unnormalized layer weight as
\begin{equation}
    \widetilde{w}_k
    =
    \sigma\left(\frac{k-s}{\tau}\right)
    \sigma\left(\frac{e-k}{\tau}\right),
    \qquad k\in\mathcal{K},
\end{equation}
where $\sigma(\cdot)$ is the sigmoid function, $[s,e]$ specifies the central intervention region, and $\tau$ controls the smoothness of the transition toward its boundaries. This produces larger weights within the central region and smoothly attenuates the intervention strength toward shallower and deeper layers.

To keep the overall intervention scale comparable across architectures and layer ranges, we apply RMS normalization:
\begin{equation}
    w_k
    =
    \frac{\widetilde{w}_k}
    {
        \sqrt{
            |\mathcal{K}|^{-1}
            \sum_{j\in\mathcal{K}}
            \widetilde{w}_j^2
        }
    }.
\end{equation}
The normalized weights therefore satisfy
\begin{equation}
    \frac{1}{|\mathcal{K}|}
    \sum_{k\in\mathcal{K}}
    w_k^2
    =
    1,
\end{equation}
so that the language-specific perturbation ratio $\rho_{\ell}$ retains a comparable layer-averaged interpretation even when the number and depth of intervened layers differ across model architectures.

We use
$(\mathcal{K},s,e,\tau)=(12\text{--}32,14,27,1.5)$
for InternVL3.5-14B-IT (40 layers),
$(11\text{--}29,13,24,1.35)$
for Qwen3-VL-8B-IT, and
$(14\text{--}39,17,33,1.8)$
for Gemma-3-12B-IT.
These ranges are transferred according to normalized decoder depth rather than tuned independently on downstream evaluation sets.

\section{Calibration of the Perturbation Ratio}
\label{app:rho_calibration}

The appropriate intervention strength varies across languages because the
magnitude and orientation of the non-shared residual depend on the underlying
representation geometry. We therefore determine a language-specific
perturbation ratio in two stages. A geometry-based estimate
$\hat{\rho}_{\ell}$ first provides an initial search scale, after which a
short search on a dedicated calibration set refines it to the final ratio
$\rho_{\ell}^{\star}$. Calibration is performed separately for each backbone,
and all ratios are frozen before evaluation on AMBER, Object HalBench, and
LLaVA-Bench.

\paragraph{Calibration data.}
We construct the calibration set from MS COCO train2014~\citep{lin2014microsoft}.
The images are disjoint from both those used to estimate the multilingual
and visual subspaces and all downstream evaluation sets. We stratify images
by the number of annotated object categories and apply DINOv2-based
coreset selection~\citep{oquab2023dinov2} to preserve visual diversity.
The selected images are fixed in advance and divided into a 300-image
selection split and a disjoint 200-image confirmation split. Prompts are
generic image-description instructions translated into each language and
contain no questions or examples from the downstream benchmarks.

\paragraph{Geometry-based estimate.}
For sample $i$, language $\ell$, phase $p$, and layer $k$, let
$\mathbf{r}_{i,\ell,p}^{(k)}$ denote the visually protected residual defined
in Eq.~\eqref{equ:r}. Its normalized removal direction is
\begin{equation}
\mathbf{u}_{i,\ell,p}^{(k)}
=
\frac{\mathbf{r}_{i,\ell,p}^{(k)}}
     {\bigl\|\mathbf{r}_{i,\ell,p}^{(k)}\bigr\|_2+\epsilon}.
\end{equation}
These statistics are computed on the aligned fitting corpus
(Sec.~\ref{sec:method}), where each target-language instance has an English
counterpart. We characterize each language using one benefit term and two
risk terms.

The \emph{alignment benefit} measures the local increase in similarity to the
aligned English representation under a small intervention. We first define
the probe-intervened state as
\begin{equation}
\widetilde{\mathbf{h}}_{i,\ell,p}^{(k)}
=
\mathbf{h}_{i,\ell,p}^{(k)}
-
\rho_{\mathrm{probe}}
\bigl\|\mathbf{h}_{i,\ell,p}^{(k)}\bigr\|_2
\mathbf{u}_{i,\ell,p}^{(k)} .
\label{eq:probe_state}
\end{equation}
The alignment benefit is then
\begin{equation}
\begin{aligned}
A_{\ell}
=
\frac{1}{\rho_{\mathrm{probe}}}
\mathbb{E}_{i,p,k}
\Bigl[
&\cos\bigl(
\widetilde{\mathbf{h}}_{i,\ell,p}^{(k)},
\mathbf{h}_{i,\mathrm{en},p}^{(k)}
\bigr)
\\[-2pt]
&-
\cos\bigl(
\mathbf{h}_{i,\ell,p}^{(k)},
\mathbf{h}_{i,\mathrm{en},p}^{(k)}
\bigr)
\Bigr],
\end{aligned}
\label{eq:alignment_gain}
\end{equation}
where $\rho_{\mathrm{probe}}=0.001$ is shared across languages.
Equation~\eqref{eq:alignment_gain} is a finite-difference estimate of the
local alignment gain. English serves only as the aligned reference; the
intervention direction remains language-specific.

The \emph{language-preservation risk} penalizes overlap with the
language-centroid direction, which Observation~2 shows controls
output-language realization. Let
\[
\mathbf{d}_{\ell,p}^{(k)}
=
\boldsymbol{\mu}_{\ell,p}^{(k)}
-
\boldsymbol{\mu}_{\mathrm{en},p}^{(k)} .
\]
We define
\begin{equation}
D_{\ell}^{\mathrm{lang}}
=
\mathbb{E}_{i,p,k}
\left[
\frac{
\bigl(
{\mathbf{u}_{i,\ell,p}^{(k)}}^{\top}
\mathbf{d}_{\ell,p}^{(k)}
\bigr)^2
}{
\bigl\|\mathbf{d}_{\ell,p}^{(k)}\bigr\|_2^2+\epsilon
}
\right].
\label{eq:language_risk}
\end{equation}

The \emph{visual-preservation risk} measures the fraction of the removal
direction lying in the protected visual subspace of Sec.~\ref{sec:method},
with projector $P_{v,\ell,p}^{(k)}$:
\begin{equation}
D_{\ell}^{\mathrm{vis}}
=
\mathbb{E}_{i,p,k}
\left[
\bigl\|
P_{v,\ell,p}^{(k)}
\mathbf{u}_{i,\ell,p}^{(k)}
\bigr\|_2^2
\right].
\label{eq:visual_risk}
\end{equation}

In all three terms, samples are weighted uniformly, the prefill and decoding
phases receive equal weight, and layers within each phase are weighted by
$w_k^2$ according to their contribution to the total perturbation energy. We combine the three terms into the
dimensionless score
\begin{equation}
S_{\ell}
=
\frac{
\max(A_{\ell},0)
}{
\sqrt{
D_{\ell}^{\mathrm{lang}}
+
D_{\ell}^{\mathrm{vis}}
}
+\epsilon
},
\label{eq:geometry_score}
\end{equation}
and map it to the intervention scale as
\begin{equation}
\hat{\rho}_{\ell}
=
\operatorname{clip}
\bigl(
\lambda S_{\ell},
\,0,\,
\rho_{\max}
\bigr),
\qquad
\rho_{\max}=0.02 .
\label{eq:geometry_rho}
\end{equation}
The two risk terms are weighted equally. Because the effective intervention
scale differs substantially across backbone geometries, we use fixed
backbone-specific scale factors $\lambda=0.025$, $0.15$, and $0.25$ for
Gemma-3-12B-IT, Qwen3-VL-8B-IT, and InternVL3.5-14B-IT, respectively. These
values only initialize the local search; the final ratio is determined
independently for each language by the calibration procedure.
Equation~\eqref{eq:geometry_rho} is therefore an initialization rule rather than an analytic optimum: it favors larger ratios when the alignment gain is high
and overlap with language-control and visually sensitive directions is low.

\paragraph{Calibration criterion.}
For each candidate ratio, we generate responses on the selection split using
the same prompts, random seed, and decoding configuration as the paired
baseline. Generated object mentions are mapped to canonical COCO categories
by GPT-5.5, which is used only for extraction and not for correctness
judgment. The extracted objects are then compared deterministically against
a fixed ground-truth inventory formed by the union of instance-segmentation
labels and objects mentioned in the five human COCO captions.

We compute CHAIR$_s$, CHAIR$_i$, object recall, and target-language adherence.
A candidate is admissible if, relative to the paired baseline,
(i) neither CHAIR$_s$ nor CHAIR$_i$ increases,
(ii) object recall decreases by at most $1.5\%$ in relative terms, and
(iii) target-language adherence decreases by at most $2$ absolute percentage
points. The recall constraint prevents reductions in hallucination obtained
simply by suppressing valid object mentions.

Among admissible candidates, we maximize the mean relative reduction in
CHAIR$_s$ and CHAIR$_i$:
\begin{equation}
\begin{aligned}
\mathcal{C}_{\ell}(\rho)
={}&
\frac{
\mathrm{CHAIR}_s^{\mathrm{base}}
-
\mathrm{CHAIR}_s(\rho)
}{
2\,\mathrm{CHAIR}_s^{\mathrm{base}}
}
\\[-2pt]
&+
\frac{
\mathrm{CHAIR}_i^{\mathrm{base}}
-
\mathrm{CHAIR}_i(\rho)
}{
2\,\mathrm{CHAIR}_i^{\mathrm{base}}
}.
\end{aligned}
\label{eq:calibration_score}
\end{equation}
Inadmissible candidates are excluded from the search. If no candidate is
admissible, we set $\rho_{\ell}^{\star}=0$ and apply no intervention. This
fallback is not triggered for any reported model--language pair.

\paragraph{Local search.}
The search is performed independently for each language in
$x=\log_{10}\rho$ space. Starting from $\hat{\rho}_{\ell}$, we evaluate
\begin{equation}
\left\{
\frac{\hat{\rho}_{\ell}}{\kappa},
\;
\hat{\rho}_{\ell},
\;
\kappa\hat{\rho}_{\ell}
\right\},
\qquad
\kappa=2.5.
\label{eq:rho_bracket}
\end{equation}
If an endpoint outperforms the center, the bracket is shifted by a factor of
$\kappa$ in that direction, requiring only one new evaluation. Shifting
continues until the center scores highest or $\rho_{\max}$ is
reached~\citep{press1989numerical}.

Let $x_0$ denote the center and $C_{-}$, $C_0$, and $C_{+}$ the scores of the
three bracket points. Following the parabolic step of Brent's
method~\citep{brent2013algorithms}, we take the vertex of the quadratic
interpolant:
\begin{equation}
x^{\dagger}
=
x_0
+
\frac{\log_{10}\kappa}{2}
\frac{
C_{-}-C_{+}
}{
C_{-}-2C_0+C_{+}
}.
\label{eq:parabolic_step}
\end{equation}
When the quadratic fit is degenerate or numerically ill-conditioned, we
retain the bracket center, i.e., $x^{\dagger}=x_0$. Otherwise, because the
center is locally best, the interpolated estimate lies within half a search
step of $x_0$.

\paragraph{Rounding and confirmation.}
We round $10^{x^{\dagger}}$ to the nearest point in $\log_{10}\rho$ space on
the grid
\[
\mathcal{G}
=
\{1,\,1.5,\,2.5,\,4,\,5\}
\times 10^k,
\qquad
k\in\mathbb{Z},
\]
and use the resulting value as the candidate $\rho_{\ell}^{\star}$. The grid
is used only to report the ratio at limited precision. Its points are
approximately uniformly spaced in log space, and rounding changes the
interpolated value by at most a factor of $\sqrt{2}$, finer than the search
step $\kappa$.

Because the rounded value need not have been evaluated during the search, we
evaluate it once on the disjoint confirmation split using the same
admissibility criteria. Only a candidate that remains admissible is frozen as
$\rho_{\ell}^{\star}$. All reported model--language ratios pass this
confirmation check. Table~\ref{tab:target_rho} reports the final ratios,
which are fixed before any downstream evaluation.

\section{Evaluation Metrics}
\label{app:evaluation_metrics}
We evaluate hallucination, visual coverage, and response quality using complementary metrics. On AMBER, Cover measures the fraction of annotated objects mentioned in a response, while CHAIR measures the fraction of extracted objects unsupported by the image. Hal measures the proportion of responses containing at least one hallucinated object, and Cog measures the fraction of extracted objects matching the annotated hallucinatory targets. On Object HalBench, $\mathrm{CHAIR}_s$ and $\mathrm{CHAIR}_i$ measure response-level and object-level hallucination rates, respectively, while Recall measures the fraction of ground-truth objects recovered. On LLaVA-Bench, we report visual accuracy, hallucination-free quality, coverage, relevance, target-language adherence, and overall response quality, together with win/tie/loss counts against the paired baseline. Lower values are better for hallucination metrics, whereas higher values are better for coverage, recall, and quality metrics. All scores are reported on a 0--100 scale, with main results macro-averaged across the ten target languages.

\section{Accumulation of Small Representation Corrections}
\label{app:butterfly_derivation}

This section provides the detailed derivation underlying the
trajectory-level accumulation discussed in the main text.
We show that although each MVP-Steer update is locally small relative
to the current hidden state, repeatedly applying such structured
corrections across layers and autoregressive decoding steps can yield
a non-negligible sequence-level intervention.

\paragraph{Local perturbation magnitude.}
Let $\mathbf{h}^{(k)}_{\ell,t}\in\mathbb{R}^{d}$ denote the hidden state
for language $\ell$ at decoding step $t$ and intervention layer $k$, and
let $\Delta\mathbf{h}^{(k)}_{\ell,t}$ denote the update introduced by
MVP-Steer. From the norm-scaled intervention defined in
Section~\ref{sec:method_inference}, the update is
\begin{equation}
    \Delta\mathbf{h}^{(k)}_{\ell,t}
    =
    -a^{(k)}_{\ell,t}
    \mathbf{r}^{(k)}_{\gamma,\ell,t},
\end{equation}
where
\begin{equation}
    a^{(k)}_{\ell,t}
    =
    \min\left(
        a_{\max},
        \rho_{\ell} w_k
        \frac{
            \left\|\mathbf{h}^{(k)}_{\ell,t}\right\|_2
        }{
            \left\|\mathbf{r}^{(k)}_{\gamma,\ell,t}\right\|_2+\epsilon
        }
    \right).
\end{equation}
Here, $\rho_{\ell}$ is the language-specific perturbation strength and
$w_k$ is the depth-dependent layer weight.

When clipping is inactive,
\begin{equation}
    \frac{
        \left\|\Delta\mathbf{h}^{(k)}_{\ell,t}\right\|_2
    }{
        \left\|\mathbf{h}^{(k)}_{\ell,t}\right\|_2
    }
    =
    \rho_{\ell}w_k
    \frac{
        \left\|\mathbf{r}^{(k)}_{\gamma,\ell,t}\right\|_2
    }{
        \left\|\mathbf{r}^{(k)}_{\gamma,\ell,t}\right\|_2+\epsilon
    }
    \approx
    \rho_{\ell}w_k ,
\end{equation}
where the approximation holds when the residual norm is large relative
to the numerical stabilizer $\epsilon$.
When clipping is active, the realized update is smaller than this
nominal target. Thus, in general,
\begin{equation}
    \frac{
        \left\|\Delta\mathbf{h}^{(k)}_{\ell,t}\right\|_2
    }{
        \left\|\mathbf{h}^{(k)}_{\ell,t}\right\|_2
    }
    \leq
    \rho_{\ell}w_k .
\end{equation}

As described in Section~\ref{sec:method_inference}, the layer weights are
RMS-normalized over the intervention-layer set $\mathcal{K}$:
\begin{equation}
    \frac{1}{|\mathcal{K}|}
    \sum_{k\in\mathcal{K}} w_k^2
    = 1.
\end{equation}
The exact weighting function and architecture-specific layer ranges are
given in Appendix~\ref{app:layer_weight}.

\paragraph{Accumulated perturbation energy.}
We next quantify the aggregate scale of these locally small updates.
For a decoding trajectory of $T$ steps, define the total perturbation
energy across all intervention layers as
\begin{equation}
    E_{\Delta}
    =
    \sum_{t=1}^{T}
    \sum_{k\in\mathcal{K}}
    \left\|
        \Delta\mathbf{h}^{(k)}_{\ell,t}
    \right\|_2^2 .
\end{equation}
Using the local relative-perturbation bound above gives
\begin{equation}
    E_{\Delta}
    \leq
    \rho_{\ell}^{2}
    \sum_{t=1}^{T}
    \sum_{k\in\mathcal{K}}
    w_k^2
    \left\|
        \mathbf{h}^{(k)}_{\ell,t}
    \right\|_2^2 .
\end{equation}
In the typical unclipped regime, the inequality is approximately tight
up to the numerical stabilizer.

To expose the scaling behavior, we approximate the hidden-state energy
as roughly proportional to the hidden dimension,
\begin{equation}
    \left\|
        \mathbf{h}^{(k)}_{\ell,t}
    \right\|_2^2
    \approx d .
\end{equation}
Together with the RMS normalization
$\sum_{k\in\mathcal{K}}w_k^2=|\mathcal{K}|$, this yields
\begin{equation}
    E_{\Delta}
    \approx
    T|\mathcal{K}|d\rho_{\ell}^{2}.
    \label{eq:accumulated_energy}
\end{equation}
Therefore, even when the correction at each individual token--layer
position is small, its aggregate energy grows with the generation
length $T$, the number of intervened layers $|\mathcal{K}|$, and the
hidden dimension $d$.

For simplicity, Eq.~\ref{eq:accumulated_energy} considers only
autoregressive decoding states. MVP-Steer additionally intervenes on
the final prefill state before answer generation, as specified in
Section~\ref{sec:method_inference}. Including this state adds one further
set of layer-wise corrections and does not alter the scaling argument.

\paragraph{Propagation through the autoregressive trajectory.}
Accumulated perturbation energy characterizes the total scale of the
local interventions, but does not by itself determine their effect on a
later model state. Earlier corrections are propagated through subsequent
transformer computation and autoregressive feedback.

Consider a later hidden representation
$\mathbf{h}^{(k')}_{\ell,T}$ at decoding step $T$ and layer $k'$.
For a locally fixed generation trajectory, let
\begin{equation}
    \mathbf{J}_{T,k'\leftarrow t,k}
    =
    \frac{
        \partial \mathbf{h}^{(k')}_{\ell,T}
    }{
        \partial \mathbf{h}^{(k)}_{\ell,t}
    }
\end{equation}
denote the Jacobian mapping a perturbation introduced at decoding step
$t$ and layer $k$ to its downstream effect on the later representation.
Under a first-order approximation,
\begin{equation}
    \Delta\mathbf{h}^{(k')}_{\ell,T}
    \approx
    \sum_{t\leq T}
    \sum_{k\in\mathcal{K}}
    \mathbf{J}_{T,k'\leftarrow t,k}
    \Delta\mathbf{h}^{(k)}_{\ell,t}.
    \label{eq:trajectory_propagation}
\end{equation}

For compactness, index each token--layer correction by $i$ and define
\begin{equation}
    \mathbf{z}_i
    =
    \mathbf{J}_i\Delta\mathbf{h}_i .
\end{equation}
The squared magnitude of the resulting downstream change can then be
expanded as
\begin{equation}
    \left\|
        \Delta\mathbf{h}^{(k')}_{\ell,T}
    \right\|_2^2
    \approx
    \sum_i \|\mathbf{z}_i\|_2^2
    +
    2\sum_{i<j}
    \left\langle
        \mathbf{z}_i,
        \mathbf{z}_j
    \right\rangle .
    \label{eq:propagation_cross_terms}
\end{equation}
The first term captures the accumulated downstream contribution of
individual corrections, whereas the second captures interactions among
corrections introduced at different token--layer positions.
These cross terms need not always be positive: propagated updates may
reinforce or partially cancel one another.

MVP-Steer, however, does not inject independent random perturbations.
At each generation step, it repeatedly attenuates a structured
non-shared residual defined by the same fitted representation
decomposition. Consequently, successive corrections can exert coherent
downstream effects on the evolving generation trajectory rather than
behaving as unrelated noise.

Eqs.~\eqref{eq:accumulated_energy} and~\eqref{eq:trajectory_propagation} do not imply that every local perturbation is amplified or that effects accumulate monotonically. They establish two properties:
individually small interventions can have a non-negligible aggregate
scale when repeatedly applied, and earlier interventions can influence
later states through autoregressive propagation.
Together, these properties provide a mechanism by which small,
structured representation-level corrections can produce substantially
larger sequence-level changes in visual grounding and hallucination
behavior. We refer to this cumulative mechanism as the
\emph{butterfly effect} of language-specific components.

\section{Direction Specificity under a Matched Perturbation Budget}
\label{app:Effect of the Steering Direction}
To test whether the gains of MVP-Steer come from the decomposed direction or from
small perturbations in general, we keep the per-state perturbation budget of MVP-Steer
fixed and change only the direction. At each intervened state, we apply
$\mathbf{h}'=\mathbf{h}-q\,\mathbf{d}/(\|\mathbf{d}\|_2+\epsilon)$, where $q$ is the
norm of the MVP-Steer update on the same state. We compare four directions: a fixed
random direction in the shared-space complement, drawn once per language, phase and
layer and shared across all tokens; a dynamic random direction resampled at every
token and layer; the shared component $\mathbf{s}$; and the visual component
$\mathbf{v}$. Both random directions use the same soft visual protection as
MVP-Steer. All other settings, including $\rho_\ell$, layer weights, decoding
configuration and evaluation data, follow the main experiment.

\begin{table*}[t]
\centering
\small
\setlength{\tabcolsep}{5pt}
\begin{tabular}{llcccc}
\toprule
Direction & $\mathbf{d}$ & Cover$\uparrow$ & CHAIR$\downarrow$ & Hal$\downarrow$ & Cog$\downarrow$ \\
\midrule
Baseline & -- & 48.47 & 4.52 & 12.98 & 2.84 \\
\midrule
Fixed random & $(\mathbf{I}-\gamma\mathbf{P}_v)(\mathbf{I}-\mathbf{P}_s)\,\mathbf{g}_{\ell,p,k}$
  & \textbf{48.56} (+0.20) & 4.74 ($-$4.69) & 13.91 ($-$7.17) & 3.00 ($-$5.63) \\
Dynamic random & $(\mathbf{I}-\gamma\mathbf{P}_v)(\mathbf{I}-\mathbf{P}_s)\,\mathbf{g}_{i,p,k,t}$
  & 48.26 ($-$0.43) & 4.40 (+2.65) & 13.04 ($-$0.48) & 2.82 (+0.67) \\
Shared-only & $\mathbf{s}$
  & 48.22 ($-$0.51) & 4.64 ($-$2.61) & 13.34 ($-$2.79) & 2.90 ($-$2.11) \\
Visual-only & $\mathbf{v}$
  & 48.31 ($-$0.32) & 4.40 (+2.72) & 13.20 ($-$1.67) & 2.85 ($-$0.32) \\
\midrule
MVP-Steer & $\mathbf{r}_\gamma$
  & 48.44 ($-$0.06) & \textbf{3.90} (+13.77) & \textbf{11.34} (+12.60) & \textbf{2.51} (+11.79) \\
\bottomrule
\end{tabular}
\caption{Direction controls on InternVL3.5-14B-IT (AMBER, averaged over ten languages).
All conditions use the per-state perturbation norm of MVP-Steer on the same state,
$q=\min(\rho_\ell w_k\|\mathbf{h}\|_2,\ a_{\max}\|\mathbf{r}_\gamma\|_2)$, with the same
$\rho_\ell$, layer weights and decoding configuration; only the direction $\mathbf{d}$
differs. $\mathbf{g}\sim\mathcal{N}(\mathbf{0},\mathbf{I})$ is drawn once per language,
phase and layer for the fixed control, and resampled at every sample, token and layer
for the dynamic control. Numbers in parentheses are relative improvements (\%) over the
baseline. Positive values indicate improvement for all metrics.}
\label{tab:direction_controls}
\end{table*}

Table~\ref{tab:direction_controls} compares MVP-Steer with alternative intervention directions. The fixed random direction increases all three hallucination metrics, raising CHAIR by 4.7\% and Hal by 7.2\% relative to the baseline. Attenuating the shared component also increases all three metrics, supporting the preservation of shared semantics. The dynamic random and visual-only directions reduce CHAIR by at most 2.7\% and do not reduce Hal, whereas MVP-Steer reduces CHAIR by 13.8\% and Hal by 12.6\%. Coverage stays within 0.6\% of the baseline in all conditions. These results highlight the effectiveness of the decomposition-derived residual direction in reducing hallucination while maintaining coverage.

\section{Detailed Layer-Wise Cross-Lingual Alignment Analysis}
\label{app:Detailed layer-wise cross-lingual alignment analysis}
\begin{figure}[t]
    \centering
    \includegraphics[width=\linewidth]{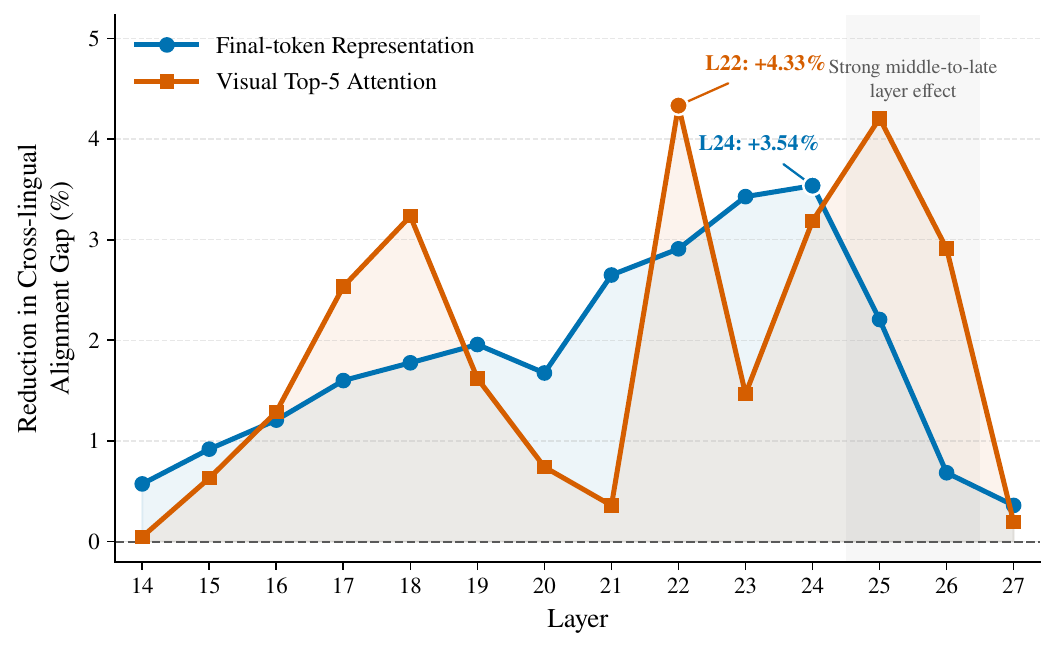}
    \caption{
    Layer-wise mechanistic effect of MVP-Steer on InternVL3.5-14B-IT across the primary
    intervention layers (L14--27).
    We report the relative improvement in cross-lingual cosine similarity for
    final-token hidden representations and the visual attention distributions of
    the Top-5 visually sensitive heads.
    }
    \label{fig:layerwise_alignment}
\end{figure}

We provide additional details for the mechanistic analysis presented in Figure~\ref{fig:layerwise_alignment}. All experiments are conducted on InternVL3.5-14B-IT and cover ten target languages, resulting in \(\binom{10}{2}=45\) unordered cross-lingual language pairs for each sample. All intervention parameters and visual-head selections are fixed prior to the alignment evaluation.

\textbf{Representation alignment.}
We use 100 fixed images paired with semantically equivalent target responses in the ten languages. Each target response is processed using teacher forcing so that the comparison isolates differences in internal representations under matched semantic content. At each transformer layer, we extract the hidden state corresponding to the final answer token and compute cosine similarity between every language pair associated with the same image and semantic content. The similarities are subsequently aggregated over the 45 language pairs and all evaluation images. Using the final answer token provides a controlled representation after the model has processed the complete target response while avoiding averaging effects caused by differences in sequence length and token position across languages.

\textbf{Visual-attention alignment.}
We separately evaluate whether MVP-Steer also makes different languages rely on more consistent visual evidence. Using 100 held-out images, we construct semantically equivalent multilingual queries for the same visual input. At each layer, we extract the attention from the final prompt token to the image tokens and normalize the attention weights over the visual-token dimension before cross-lingual comparison. To focus the analysis on visually informative heads, we identify visually sensitive heads using an independent calibration set of 100 COCO images. Attention heads are ranked according to how strongly their behavior distinguishes correctly matched images from mismatched images, and this ranking is frozen before the multilingual alignment evaluation. The main analysis uses the Top-5 visually sensitive heads in each layer. Cross-lingual cosine similarity is then computed over the normalized visual attention distributions and aggregated across samples and language pairs. Consequently, this metric captures whether different languages attend to more similar visual locations.

For example, at L24, final-token cross-lingual similarity increases from \(0.849658\) to \(0.854975\). This absolute cosine increase of \(0.005317\) corresponds to removing \(3.54\%\) of the remaining gap to perfect alignment. Similarly, for the Top-5 visually sensitive heads at L22, similarity increases from \(0.937235\) to \(0.939954\), corresponding to a \(4.33\%\) reduction in the remaining visual-attention alignment gap.

\textbf{Full layer-wise results.}
Table~\ref{tab:layerwise_alignment_full} reports the complete results over L12--32. We distinguish the primary intervention range L14--27 from L12--13 and L28--32, which mainly serve as smoothing shoulders of the layer-wise intervention schedule. Within L14--27, final-token representation alignment improves at all 14 layers, with an average alignment-gap reduction of \(1.82\%\) and a maximum of \(3.54\%\) at L24. Visual Top-5 attention alignment likewise improves at all 14 layers, with an average gap reduction of \(1.91\%\) and the largest improvements at L22 (\(4.33\%\)) and L25 (\(4.21\%\)). The concentration of alignment gains within the primary intervention range further connects MVP-Steer's layer-wise intervention to stronger cross-lingual consistency in both internal representations and visual attention.

\begin{table*}[t]
\centering
\small

\setlength{\tabcolsep}{3.4pt}
\renewcommand{\arraystretch}{1.10}

\begin{tabular}{@{}lcccccccc@{}}
\toprule

\multirow{2}{*}{\textbf{Layer}}
&
\multicolumn{4}{c}{\textbf{Final-token Representation}}
&
\multicolumn{4}{c}{\textbf{Visual Top-5 Attention}}
\\

\cmidrule(lr){2-5}
\cmidrule(lr){6-9}

&
\textbf{Base}
&
\textbf{+MVP-Steer}
&
\(\boldsymbol{\Delta}\)
&
\textbf{GapRed.}
&
\textbf{Base}
&
\textbf{+MVP-Steer}
&
\(\boldsymbol{\Delta}\)
&
\textbf{GapRed.}
\\

\midrule

L12
& 0.796568 & 0.796850 & +0.000282 & +0.14\%
& 0.958802 & 0.958802 & +0.000000 & +0.00\% \\

L13
& 0.797245 & 0.797922 & +0.000677 & +0.33\%
& 0.964243 & 0.964347 & +0.000104 & +0.29\% \\

\midrule

L14
& 0.802769 & 0.803900 & +0.001131 & +0.57\%
& 0.897084 & 0.897133 & +0.000049 & +0.05\% \\

L15
& 0.805917 & 0.807701 & +0.001784 & +0.92\%
& 0.920083 & 0.920587 & +0.000504 & +0.63\% \\

L16
& 0.808169 & 0.810485 & +0.002316 & +1.21\%
& 0.891346 & 0.892747 & +0.001401 & +1.29\% \\

L17
& 0.811737 & 0.814750 & +0.003013 & +1.60\%
& 0.921826 & 0.923807 & +0.001981 & +2.53\% \\

L18
& 0.813202 & 0.816521 & +0.003319 & +1.78\%
& 0.929576 & 0.931853 & +0.002277 & +3.23\% \\

L19
& 0.800787 & 0.804688 & +0.003901 & +1.96\%
& 0.921586 & 0.922858 & +0.001272 & +1.62\% \\

L20
& 0.791704 & 0.795194 & +0.003490 & +1.68\%
& 0.955741 & 0.956069 & +0.000328 & +0.74\% \\

L21
& 0.816378 & 0.821242 & +0.004864 & +2.65\%
& 0.951337 & 0.951512 & +0.000175 & +0.36\% \\

L22
& 0.819559 & 0.824808 & +0.005249 & +2.91\%
& 0.937235 & 0.939954 & +0.002719 & \textbf{+4.33\%} \\

L23
& 0.827550 & 0.833461 & +0.005911 & +3.43\%
& 0.936872 & 0.937798 & +0.000926 & +1.47\% \\

L24
& 0.849658 & 0.854975 & +0.005317 & \textbf{+3.54\%}
& 0.945541 & 0.947276 & +0.001735 & +3.19\% \\

L25
& 0.844988 & 0.848409 & +0.003421 & +2.21\%
& 0.874996 & 0.880254 & +0.005258 & +4.21\% \\

L26
& 0.806424 & 0.807749 & +0.001325 & +0.68\%
& 0.836307 & 0.841071 & +0.004764 & +2.91\% \\

L27
& 0.782173 & 0.782957 & +0.000784 & +0.36\%
& 0.957577 & 0.957659 & +0.000082 & +0.19\% \\

\midrule

L28
& 0.770044 & 0.770508 & +0.000464 & +0.20\%
& 0.949110 & 0.949684 & +0.000574 & +1.13\% \\

L29
& 0.727170 & 0.726662 & -0.000508 & -0.19\%
& 0.943309 & 0.943291 & -0.000018 & -0.03\% \\

L30
& 0.701133 & 0.700928 & -0.000205 & -0.07\%
& 0.953280 & 0.953654 & +0.000374 & +0.80\% \\

L31
& 0.665726 & 0.665618 & -0.000108 & -0.03\%
& 0.903085 & 0.903603 & +0.000518 & +0.53\% \\

L32
& 0.612733 & 0.613217 & +0.000484 & +0.12\%
& 0.950576 & 0.950526 & -0.000050 & -0.10\% \\

\bottomrule
\end{tabular}
\caption{
Full layer-wise cross-lingual alignment results on InternVL3.5-14B-IT.
For both final-token representations and the Top-5 visually sensitive attention heads,
we report cosine similarity before and after MVP-Steer, the absolute change
$\Delta$, and the reduction in alignment gap (\textsc{GapRed.}).
\textsc{GapRed.} is defined as
$(s_{\mathrm{MVP-Steer}}-s_{\mathrm{base}})/(1-s_{\mathrm{base}})\times100$.
L14--27 denotes the primary intervention range, while L12--13 and L28--32
mainly serve as smoothing layers.
}

\label{tab:layerwise_alignment_full}

\end{table*}

\end{document}